\documentclass[journal]{IEEEtran}
\ifCLASSINFOpdf
\else
\fi
\usepackage{graphicx}
\usepackage{algpseudocode}
\usepackage{algorithm}
\usepackage{times}
\usepackage{bbm}
\usepackage{epsfig}
\usepackage{times}
\usepackage{xcolor}
\usepackage{url}
\usepackage{amsmath,amssymb}
\usepackage{algorithmicx}
\usepackage{algpseudocode}
\usepackage{multirow}
\usepackage{soul}
\usepackage{xspace}
\makeatletter
\DeclareRobustCommand\onedot{\futurelet\@let@token\@onedot}
\def\@onedot{\ifx\@let@token.\else.\null\fi\xspace}

\makeatother

\begin{document}
%
\title{Embedding Visual Hierarchy with Deep Networks for Large-Scale Visual Recognition}
%
%
%

\author{Tianyi~Zhao, ~Baopeng~Zhang, ~Wei~Zhang, ~Ning~Zhou, ~Jun~ Yu, ~Jianping~Fan
\IEEEcompsocitemizethanks{\IEEEcompsocthanksitem Tianyi Zhao, Bapeng Zhang, Wei Zhang, Ning Zhou, Jun Yu and Jianping Fan are with the Department of Computer Science, University of North Carolina, Charlotte, NC 28223, USA. e-mail: jfan@uncc.edu}
\thanks{This research is supported by National Science Foundation under 1651166-CNS.}
\thanks{Copyright (c) 2016 IEEE. Personal use of this material is permitted.
        However, permission to use this material for any other purposes must
        be obtained from the IEEE by sending an email to
        pubs-permissions@ieee.org.}
}
%
%

\markboth{\bf\em IEEE TRANSACTIONS ON Image Processing, 2017}%
{Zhao \MakeLowercase{\textit{et al.}}: Embedding Visual Hierarchy with Deep Networks for Large-Scale Visual Recognition}
%



\maketitle

\begin{abstract}
In this paper, a level-wise mixture model (LMM) is developed by embedding visual hierarchy with deep networks to support large-scale visual recognition (i.e., recognizing thousands or even tens of thousands of object classes), and a Bayesian approach is used to adapt a pre-trained visual hierarchy automatically to the improvements of deep features (that are used for image and object class representation) when more representative deep networks are learned along the time.  Our LMM model can provide an end-to-end approach for jointly learning: (a) the deep networks to extract more discriminative deep features for image and object class representation; (b) the tree classifier for recognizing large numbers of object classes hierarchically; and (c) the visual hierarchy adaptation for achieving more accurate indexing of large numbers of object classes  hierarchically. By supporting joint learning of the tree classifier, the deep networks and the visual hierarchy adaptation, our LMM algorithm can provide an effective approach for controlling inter-level error propagation effectively, thus it can achieve better accuracy rates on large-scale visual recognition. Our experiments are carried on ImageNet1K and ImageNet10K image sets, and our LMM algorithm can achieve very competitive results on both the accuracy rates and the computation efficiency as compared with the baseline methods.
\end{abstract}


%
\IEEEpeerreviewmaketitle

\section{Introduction}
\IEEEPARstart{B}y breaking the complex issue of feature learning into a set of small tasks hierarchically, deep learning \cite{NIPS2012_4824,vgg2015,googlenet2015,resnet2015,sun2016} has demonstrated a divide-and-conquer process to learn more discriminative representations for large-scale visual recognition application:  each neuron on the deep networks handles one small piece of the complex task for feature learning, and all these neurons can seamlessly collaborate to accomplish the complex task for feature learning in a coarse-to-fine fashion. For large-scale visual recognition application (i.e., recognizing thousands or even tens of thousands of object classes) \cite{akata:hal-00835810,40814,Deng:2010:CMI:1888150.1888157,conf/cvpr/ZhaoX13,DengKrauseBergFei-Fei_CVPR2012,Liff,conf/eccv/DengDJFMBLNA14,LMLF}, the deep networks are usually trained in a supervised way by minimizing a flat loss function (such as cross-entropy). Some researchers have found that the neurons on the earlier layers of the deep networks are more 'common' but the neurons on the later layers are more 'specific' \cite{DBLP:journals/corr/YosinskiCBL14}. 

Even deep learning has achieved outstanding performances for many computer vision tasks, it still has room to improve: (1) strong inter-class visual similarities are typical in the domain of large-scale visual recognition especially when some object classes are fine-grained (visually-similar) \cite{conf/cvpr/SfarBG13,BirdletsFarrellICCV11,10.1109/CVPRW.2009.5206574,deng-cvpr2013}, but the $N$-way flat softmax classifier completely ignores the inter-task correlations; (2) ignoring the inter-task correlations completely may push the deep learning process away from the global optimum because the gradients of the joint objective function are not uniform for all the object classes, and such deep learning process may distract on discerning some particular object classes that are typically hard to be discriminated.

Another divide-and-conquer approach for supporting large-scale visual recognition is to leverage a pre-defined tree structure (visual hierarchy or concept ontology) \cite{imagenet_cvpr09,10.1109/MMUL.2006.63,Marszalek2007,conf/eccv/MarszalekS08, Nister:2006:SRV:1153171.1153548,conf/cvpr/GriffinP08,NIPS2010_4027,Bart:CVPR2008:TAX,NIPS2011_4212,conf/cvpr/LiuSTSL13,Sivic2008, caetano-ijcv2013,fan2008,fan2015,fan2017} to organize large numbers of object classes hierarchically. By training the tree classifier over a pre-defined tree structure hierarchically, the hierarchical visual recognition approach \cite{Dekel:2004:LMH:1015330.1015374,hierarchical-classification-via-orthogonal-transfer,Sun_ICCV2013,RePEc:bes:jnlasa:v:104:i:487:y:2009:p:1213-1223} can provide multiple advantages: (a) Making coarse-to-fine predictions along a pre-defined tree structure can effectively rule out unlikely groups of object classes (i.e., irrelevant high-level nodes on the tree structure) at an early stage, thus it can achieve sub-linear computational complexity at test time; (b) For a given group (a high-level non-leaf node on the tree structure), the learning tasks for training the inter-related classifiers for its belonging fine-grained (visually-similar) object classes are strongly inter-related, thus multi-task learning can be used to train such inter-related classifiers jointly for enhancing their discrimination power \cite{fan2008,fan2015}; (c) For a given object class, the negative images for classifier training can be selected locally from other visually-similar object classes in the same group, and the issue of huge sample imbalance can be avoided effectively; (d) For a given group, its task space for object recognition is much smaller and uniform (i.e., distinguishing only a small number of fine-grained (visually-similar) object classes in the same group rather than separating all $N$ object classes simultaneously \cite{fan2017}). 

Based on these observations, it is very nature for us to ask ourselves the following question: {\em how can we integrate these two divide-and-conquer approaches} (deep learning and hierarchical visual recognition) {\em and benefit from both of them to exploit better solutions for large-scale visual recognition?}

\begin{figure}
    \centering \includegraphics[width=\linewidth]{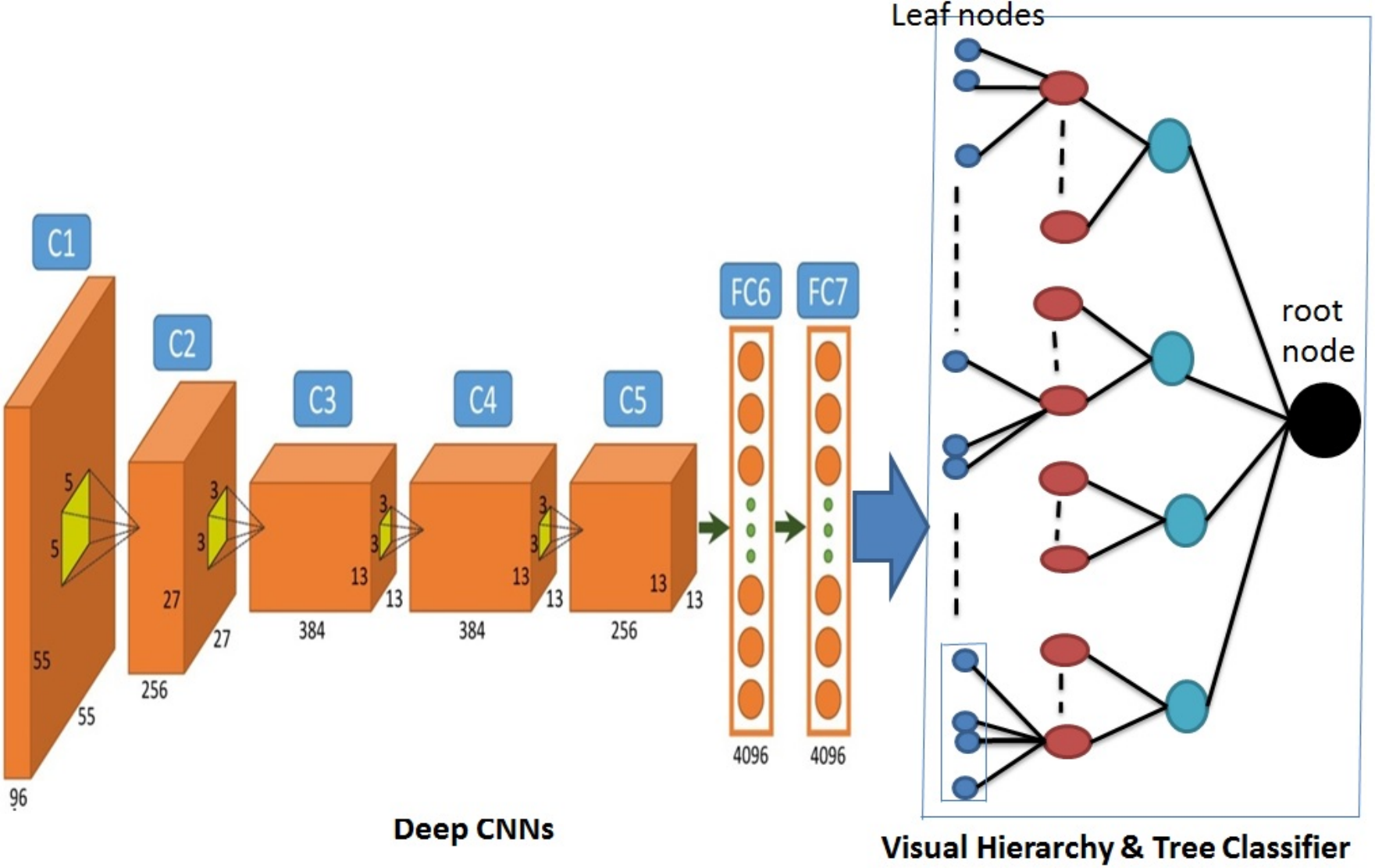}\\
    \caption{\bf The flowchart for embedding deep networks with visual hierarchy, where the tree classifier over the visual hierarchy is used to replace the traditional $N$-way flat softmax classifier.}
    \label{fig:nnp}
\end{figure}

In this paper, as shown in Fig. \ref{fig:nnp}, a level-wise mixture model (LMM) is developed by using a tree classifier to replace traditional $N$-way flat softmax classifier in the deep networks, where a visual hierarchy is embedded to: (a) organize large numbers of object classes hierarchically according to their inter-class visual similarities; (b) guide the process for joint learning of deep networks and tree classifier to make more effective splittings of the complex tasks for feature learning and hierarchical visual recognition. By leveraging group-object correlations (that are intuitively characterized by the visual hierarchy) to guide the process for jointly learning the deep networks and the tree classifier, our LMM model can boost the performance of large-scale visual recognition significantly and extract more robust and transferable features from the deep networks for image and object class representation. Because the deep features (outputs of the deep networks) are seamlessly integrated to learn the deep representations for large numbers of object classes and their inter-class visual similarities (that are used for constructing the visual hierarchy), the visual hierarchy should be adapted automatically when more representative deep networks are learned along the time, but it could be very expensive to reconstruct the visual hierarchy repeatedly. Based on this understanding, a Bayesian approach is further developed to effectively adapt the visual hierarchy during the end-to-end process for jointly learning the deep networks and the tree classifier.

In a summary, this paper has made the following {\bf\em contributions}: (1) A level-wise mixture model (LMM) is developed to embed the visual hierarchy with the deep networks, so that we can leverage the group-object (inter-level) correlations (that are intuitively characterized by the visual hierarchy) to learn more representative deep networks and more discriminative tree classifier for supporting hierarchical visual recognition; (2) A Bayesian approach is developed to adapt the visual hierarchy to the improvements of deep class representations, e.g., learning more representative deep networks along the time may result in the improvements of the deep representations for large numbers of object classes and their inter-class visual similarities. Thus our LMM model can provide an end-to-end approach for jointly learning: (a) the deep networks to extract more discriminative features for image and object class representation; (b) the tree classifier (LMM model) for recognizing large numbers of object classes hierarchically; and (c) the visual hierarchy adaptation for achieving more accurate and hierarchical indexing of large numbers of object classes and identifying the tasks with similar learning complexities automatically. By supporting joint learning of the tree classifier, the deep networks and the visual hierarchy adaptation, our LMM algorithm can provide an effective approach for controlling inter-level error propagation effectively (i.e., inter-level error propagation is a critical issue for supporting hierarchical visual recognition\cite{fan2015,fan2017}). Our proposed algorithms have achieved very competitive results on ImageNet1K and ImageNet10K image sets as compared with the baseline methods.

The rest of this paper is organized as follows. Section \ref{related work} gives a brief review of the related works on deep learning, hierarchical visual recognition and tree structures. Section \ref{LMM} introduces our level-wise mixture model (LMM) for learning the deep networks and the tree classifier jointly. Section \ref{tree structure learning} describes our algorithm for visual hierarchy construction and adaptation. Section \ref{training} provides our algorithm for learning the deep networks, the tree classifier (LMM model) and the visual hierarchy adaptation jointly in an end-to-end fashion. Our experimental results for algorithm evaluation are presented in Section \ref{Experiments}, and we conclude the paper and discuss the future works in Section \ref{Conclusion}.

\section{Related work} \label{related work}
Deep learning\cite{NIPS2012_4824,vgg2015,googlenet2015,resnet2015,sun2016} has demonstrated its outstanding abilities on learning more discriminative features and boosting the accuracy rates for large-scale visual recognition significantly. By learning more representative features and a $N$-way flat softmax classifier in an end-to-end fashion, most existing deep learning schemes have made one hidden assumption: the tasks for recognizing all the object classes are independent and share similar learning complexities. However, such assumption may not be true in many real-world applications,  e.g., strong inter-class visual similarities are typical in the domain of large-scale visual recognition especially when some object classes are fine-grained (visually-similar) \cite{conf/cvpr/SfarBG13,BirdletsFarrellICCV11,10.1109/CVPRW.2009.5206574,deng-cvpr2013}, but the $N$-way flat softmax classifier completely ignores the inter-task correlations. Ignoring the inter-task correlations completely may push the deep learning process away from the global optimum because the gradients of the joint objective function are not uniform for all the object classes, especially when they have different inter-class visual similarities and learning complexities, as a result, the deep learning process may distract on discerning some particular object classes that are typically hard to be discriminated. 

For large-scale visual recognition application (i.e., some object classes could have strong inter-class visual correlations), it is very attractive to develop new algorithms to deal with the issue of huge diversity of learning complexities more effectively, so that our deep learning schemes can effectively accomplish the task of learning more discriminative deep representations for distinguishing  visually-similar object classes effectively. A few attempts have recently been made to exploit the tree structures (both concept ontology and visual hierarchy) in the deep learning models\cite{DBLP:journals/corr/YanJDDP14,NIPS2013_5029,sun2016,peter-iccv2015,fan2017}. By integrating deep learning with multi-task learning, deep multi-task learning have received many attentions recently by using the deep networks to learn more representative features and integrating multi-task learning tools to learn inter-related classifiers jointly for separating such fine-grained (visually-similar) object classes more effectively \cite{tang-eccv2014,tan-icip2013,xue-mm-2014,share-node2014,chan-ijcv2015,wu-tip2016,zheng-wacv2014,malik-cvpr2015, darrell-cvpr2015,sun-cvpr2016}. Most existing deep multi-task learning techniques assume that all the tasks are equally related and they may completely ignore the significant differences on the inter-task relationships (inter-class visual similarities) among large numbers of object classes \cite{geman-cvpr2013,davis-iccv2011,dietterich-cvpr2009,dietterich-cvpr2009,fan2017}, e.g., some object classes may have much stronger inter-class visual similarities and not all of them have same strengths on their inter-class visual similarities. 

One intuitive way for exploiting the inter-task relationships (inter-class visual similarities) is to integrate a tree structure to organize large numbers of object classes hierarchically, e.g., the tasks for training the classifiers for the fine-grained (visually-similar) object classes under the same parent node (in the same group) may have stronger inter-task relationships and share similar learning complexities. Such tree structures can be categorized into two types: (a) concept ontology\cite{imagenet_cvpr09,10.1109/MMUL.2006.63,Marszalek2007,fan2008,caetano-ijcv2013}; and (b) label tree or visual hierarchy\cite{conf/eccv/MarszalekS08,Nister:2006:SRV:1153171.1153548,conf/cvpr/GriffinP08,NIPS2010_4027,Bart:CVPR2008:TAX,NIPS2011_4212,conf/cvpr/LiuSTSL13,Sivic2008,fan2015,fan2017}. It is worth noting that the feature space is the common space for classifier training and visual recognition \cite{fan2012}, e.g., both classifier training and visual recognition are performed in the feature space rather than in the semantic label space. Thus it could be more attractive to organize large numbers of object classes hierarchically in the feature space according to their inter-class visual correlations.

By integrating a tree structure to organize large numbers of object classes hierarchically and supervise the hierarchical process for tree classifier training, the hierarchical visual recognition approach \cite{Gao:ICCV11,Dekel:2004:LMH:1015330.1015374,hierarchical-classification-via-orthogonal-transfer,Sun_ICCV2013,RePEc:bes:jnlasa:v:104:i:487:y:2009:p:1213-1223,caetano-ijcv2013,peter-iccv2015,fan2008,fan2015} can provide many advantages, but it may seriously suffer from the problem of {\em inter-level error propagation}: the mistakes for the parent nodes will propagate to their child nodes until the leaf nodes \cite{caetano-ijcv2013,fan2015}. In addition, most existing approaches for hierarchical visual recognition focus on leveraging hand-crafted features for tree classifier training, thus it is very attractive to invest how deep features can be leveraged to improve hierarchical visual recognition \cite{peter-iccv2015,fan2017}. 

Most existing approaches for hierarchical visual recognition are static, but the process for joint learning of deep networks and tree classifier for large-scale visual recognition application is open-ended and dynamic: the deep networks for image and object class representation and the tree classifier for large-scale visual recognition may be improved along the time, e.g., more representative deep networks and more discriminative tree classifier may be achieved when more training images are added and back-propagation operations \cite{lecun1998} are continuously performed to fine-tune the weights of the deep networks. Thus most existing approaches for hierarchical visual recognition may seriously suffer from the following problem: how to adapt the pre-trained tree structure (such as visual hierarchy) to the improvements of deep networks along the time? It is worth noting that the deep networks are used to obtain the deep representations for large numbers of object classes and their inter-class visual similarities that are used for visual hierarchy construction. Thus it is very attractive to develop new approaches for jointly learning the deep networks, the tree classifier and the visual hierarchy adaptation in an end-to-end fashion.

\section{Level-wised Mixture Model (LMM)}\label{LMM}
Given $N$ object classes being recognized, when a visual hierarchy is pre-trained for organizing $N$ object classes hierarchically according to their inter-class visual similarities \cite{fan2015,fan2017}, each level of the visual hierarchy can be treated as one particular partitioning $\mathbb{T}_l $ of all these $N$ object classes (i.e., assigning $N$ object classes into a set of groups $\mathbb{T}_l $ at the $l$th level of the visual hierarchy), followed by the distribution $P$: $X \rightarrow \mathbb{T}_l$, $X$ is the deep feature space for the training image set $S$, $X = h(S,u)$, $u$ is the set of weights in the deep networks, $h(.)$ represents the mapping function of the deep networks,  $N_l$ is the number of groups at the $l$th level of the visual hierarchy, the distribution can be computed by the last layer of deep neural network, for example Softmax layer.

For a given group with the label $t$ at the $l$th level of the visual hierarchy which contains $C_t$ object classes, the probability $P(y|t)$ for each of its $C_t$ object classes can simply be defined as: $P(y|t) = 1/C_t$, e.g., we assume that the probability $P(t|l,x,w^l_t)$ for the given group $t$ (at the $l$th level of the visual hierarchy) is equally distributed among all its $C_t$ object classes. Based on this assumption, for all the groups at the $l$th level of the visual hierarchy, the distribution over $N$ object classes is defined as:
\begin{equation}\label{eq:pyl}
P(y|l, x, w^l) = \sum_t^{\mathbb{T}_l}{I(t) P(t|l,x,w^l_t) P(y|t)}
\end{equation}
where $w_l$ is the set of parameters for the node classifiers at the $l$th level of the visual hierarchy, $w^l = \{w^l_t|t \in \mathbb{T}_l\}$, $\mathbb{T}_l$ is the partitioning of $N$ object classes at the $l$th level of the visual hierarchy, each layer classifier can be treated as an additional softmax layer over the deep networks, $I(t)$ is an indication function and it is true when the object class with the label $y$ belongs to the group with the label $t$, $P(t|l,x,w^l_t)$ is the distribution of the group $t$ in one particular partitioning $\mathbb{T}_l$ at the $l$th level of the visual hierarchy  with deep representation $x$. It's worth noting that, we merging the deep neural network and the the Bayesian based Layer-wise Mixture Model by computing the probability $P(t|l,x,w^l_t)$ by deep network.

\begin{figure}
    \centering \includegraphics[width=\linewidth]{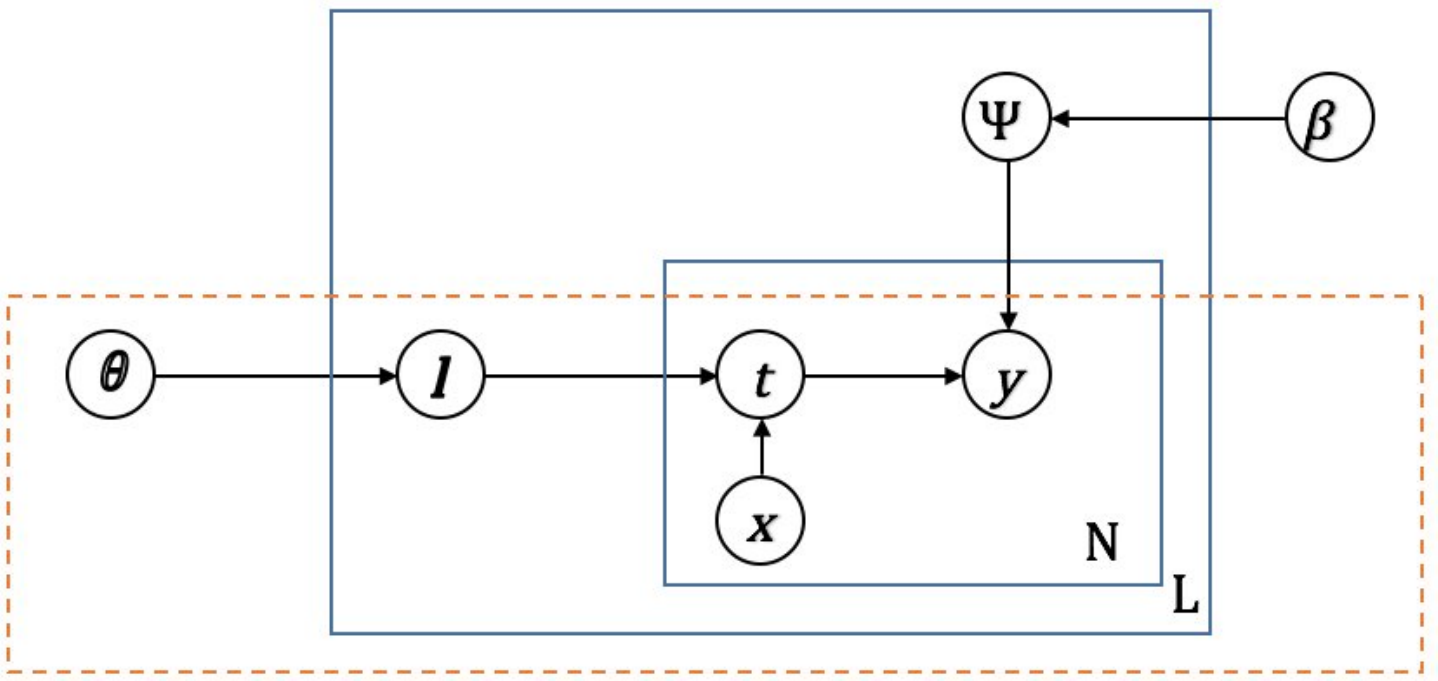}
    \caption{\bf Graph model for modeling the latent relationships between the object class $y$ and the group $t$: $l \sim Cat(\theta)$, $t \sim DL(l,x)$, $\Psi_t \sim Dir(\beta)$, $y \sim Cat(\Psi_t)$. }
    \label{fig:sample}
\end{figure}

By introducing a latent variable $\theta$ to characterize the prior distribution over all the levels of the visual hierarchy, as shown in Fig. \ref{fig:sample}, $l \sim Cat(\theta)$, $l \in \{1, \dots, L\}$, the mixture model $P(y|x,W)$ is defined for modeling the probability of the object class with the label $y$ given deep representation $x$:
\begin{equation}\label{eq:likelihood}
P(y|x,W) = \sum_{l=0}^L{\theta_l P(y|l,x,w^l)}
\end{equation}
where $W$ is the set of parameters for all the node classifiers at different levels of the visual hierarchy, $W = \{w^l|l \in \{1,\cdots, L\}\}$, $\theta_l$ is the leveraging parameter that is used to measure the contributions or effects from the node classifiers at the $l$th level of the visual hierarchy, $L$ is the depth of the visual hierarchy, e.g., the total number of levels from the root node (which contains all these $N$  object classes) to the leaf node (which contains only one particular object class).

By using our LMM model (tree classifier over the visual hierarchy) to replace traditional $N$-way flat softmax classifier, our deep networks for hierarchical visual recognition are shown in Fig. \ref{fig:nnp}. 

There are {\bf\em two significant differences} between our deep networks and traditional deep CNNs \cite{NIPS2012_4824}: (a) the tree classifier (LMM model) is used to replace the $N$-way flat softmax classifier, e.g., the tree classifier is defined as a set of node classifiers at different levels of the visual hierarchy; and (b) the group-class correlations (inter-level correlations) are leveraged to guide the process for jointly learning the deep networks and the tree classifier. Such group-object correlations (object-group assignments) are initially determined by a pre-trained visual hierarchy, and an object-group assignment matrix $\Psi$ is further learned to measure such group-object (inter-level) correlations effectively (see Section \ref{tree structure learning}). For a given group $t$ at the $l$th level of the visual hierarchy, a softmax output is used to model the probability $P(t|l,x,w^l_t)$ for the object class with the deep representation $x$ to be assigned into the given group $t$. 

After the deep networks and the tree classifier are learned jointly, for a given test image, it first goes through our deep networks to obtain its deep representation, and then such deep representation for the given test image goes through our tree classifier (our LMM model over the visual hierarchy) to obtain the prediction of its  best-matching group $t$ at each level of the visual hierarchy. The outputs from all these relevant node classifiers at different levels of the visual hierarchy are seamlessly integrated to produce the final prediction of the best-matching object class (at the bottom level of the visual hierarchy) for the given test image.
\begin{figure}
    \centering \includegraphics[width=1.0\linewidth]{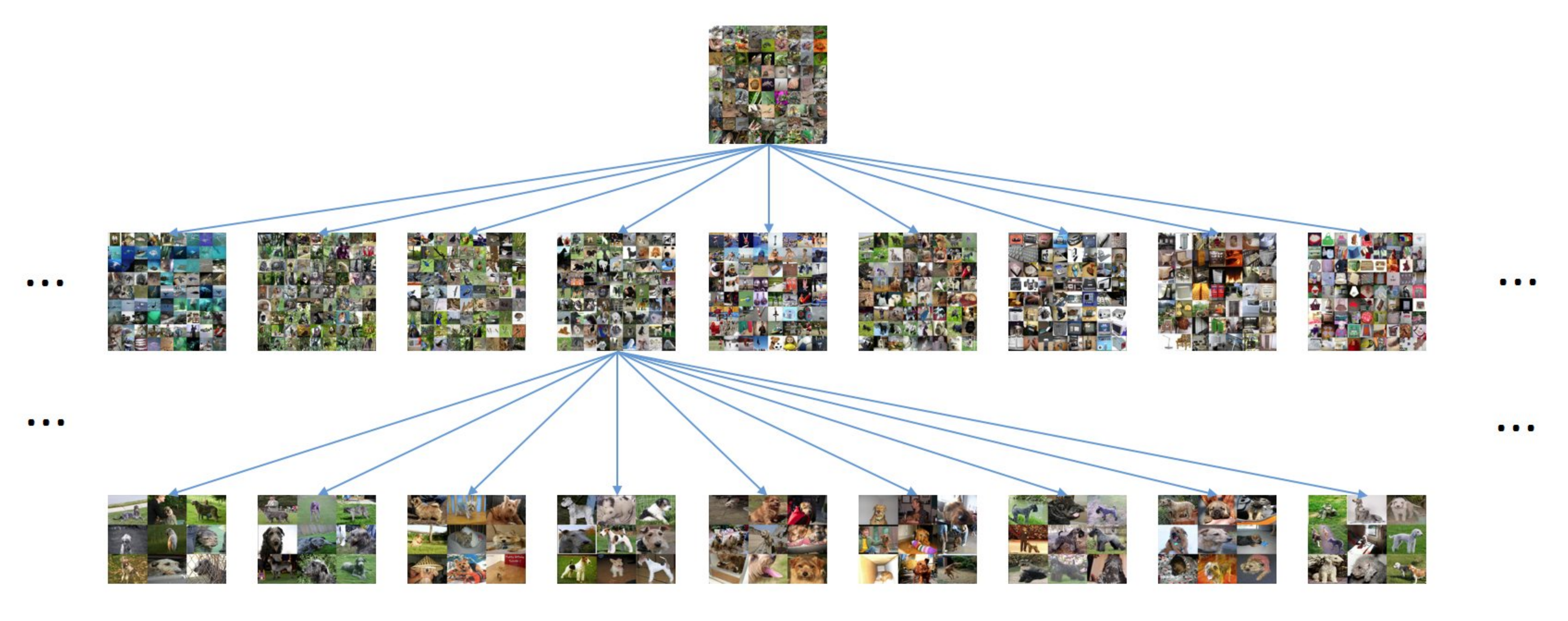}
    \caption{\bf One of our results on learning the pre-trained visual hierarchy for ImageNet1K image set.}
    \label{fig:tree1}
\end{figure}
\begin{figure}
    \centering \includegraphics[width=1.0\linewidth]{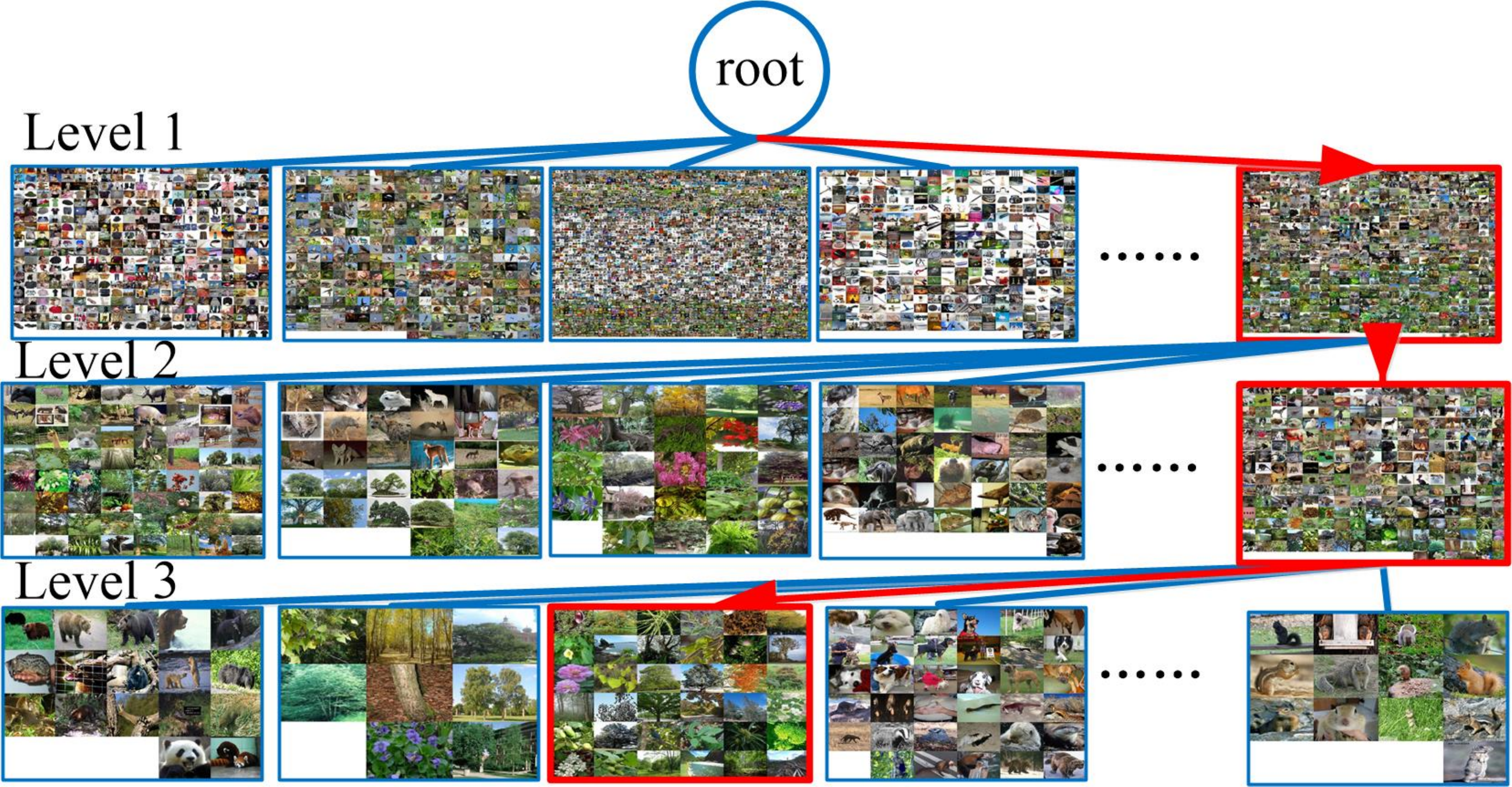}
    \caption{\bf One of our results on learning the pre-trained visual hierarchy for ImageNet10K image set.}
    \label{fig:tree2}
\end{figure}

\section{Visual Hierarchy Adaptation}\label{tree structure learning}
To learn a pre-trained visual hierarchy for organizing $N$ object classes hierarchically, we first need to calculate their inter-class visual similarities \cite{fan2015,fan2017}. For each object class, the deep networks are used to determine its deep representation from all its relevant images. The deep representation for each object class is simply obtained by averaging the deep features for all its relevant images \cite{fan2015,fan2017}. When such deep class representations are available for all these $N$ object classes, we can further calculate a $N \times N$ inter-class similarity matrix $\mathbf{S}$ and its component $S_{ij} = S(c_i, c_j)$ is defined as:
\begin{equation}\label{eq:simi}
S_{i,j} = S(c_i, c_j) = \exp \left( -\frac{d(x_i, x_j)}{\sigma} \right)
\end{equation}
where $S(c_i, c_j)$ is the inter-class visual similarity between two object classes $c_i$ and $c_j$, $d(x_i, x_j)$ is the distance between the deep class representations $x_i$ and $x_j$ for two object classes $c_i$ and $c_j$, $\sigma$ is automatically determined by a self-tune technique. When the $N \times N$ inter-class similarity matrix $\mathbf{S}$ is available, hierarchical clustering is performed to learn the pre-trained visual hierarchy \cite{fan2015,fan2017}. Two of our experimental results on learning the pre-trained visual hierarchy are shown in Fig. \ref{fig:tree1} and Fig. \ref{fig:tree2}.

In the previous section, all the object classes are treated equally, e.g., we assume that all the object classes have been assigned into the best-matching groups correctly and thus they have similar values (strengths) of the group-object correlations (i.e., all of them have good object-group assignments). In the real-world situation, it could be very hard for the pre-trained visual hierarchy to achieve optimal assignments for all the object classes because of two reasons: (a) The deep features for object class representation may not be discriminative enough because the deep networks for feature learning could be under-trained at the beginning, e.g., the deep networks may be improved along the time and the deep representations for large numbers of object classes and their inter-class visual similarities may also be improved along the time and thus the visual hierarchy should be adapted automatically (i.e., the pre-trained visual hierarchy should be adapted to the improvements of deep networks or deep class representations); (b) The hierarchical clustering technique, which is used for visual hierarchy learning, may make mistakes. 

As a result, some object classes may have stronger group-object correlations (i.e., they are strongly related with their belonging groups) because they are assigned into their best-matching groups correctly. On the other hand, some object classes may have weaker group-object correlations (i.e., they are weakly related with their belonging groups) because these object classes are assigned into some groups incorrectly. Thus it is very attractive to develop algorithms for adapting the pre-trained visual hierarchy automatically to the improvements of the outputs of the deep networks (i.e., deep features for object class representation), e.g., learning more representative deep networks along the time may result in both the improvements of deep class representations and the changes of their inter-class similarity matrix $\mathbf{S}$ for visual hierarchy construction, which may further require the changes of the group-object correlations (e.g., the underlying object-group assignments should be improved adaptively).

In this work, a Bayesian approach is developed to estimate such group-object correlations accurately and adapt the pre-trained visual hierarchy to the improvements of the group-object correlations when more representative deep networks are learned along the time. We use an object-group assignment matrix $\Psi$ to interpret the correspondences between the partitioning $\mathbb{T}_l$ of all the object classes at the $l$th level of the visual hierarchy and the labels for all its belonging object classes, $\Psi \in R^{N_l \times N}$.  $\Psi_{t, y}$ represents the probability of the object class with the label $y$ to be assigned into the given group with the label $t$, noted as $P(y|t)$. Given a pre-trained visual hierarchy, such object-group assignment matrix $\Psi$ can automatically be determined and it can be initialed as described in Section \ref{LMM}.

Our aim is to adapt the object-group assignment matrix $\Psi$ automatically to the improvements of the outputs of the deep networks along the time. The method is to sample the group label $t$ for each image, then update $\Psi$ according to the sampling results. The probability model for characterizing such object-group assignments is described in Fig. \ref{fig:sample}.

In this paper, Dirichlet-categorical model is used, $\Psi_t \sim Dir(\beta)$, $y \sim Cat(\Psi_t)$. The posterior likelihood of the group label $t$ is obtained by integrating over the object-group assignment matrix $\Psi$:
\begin{equation}\label{eq:t}
P(t_i|T^{\neg i},Y,X,l,\beta)  \propto \frac{
\Omega_{t_i,y_i}^{\neg i} + \beta_{y_i}
}{
\sum_y^\mathbb{Y} \Omega_{t_i,y}^{\neg i} + \beta_0
}
\times P(t_i|l,x_i,w^l_t)
\end{equation}
where $t_i$ is the predicted group label for the image $i$, $y_i$ is the label of the object class for the image $i$, $\mathbb{Y}$ is the set of all potential labels for the object classes, $\{1, \cdots, N\}$, $Y$ is the set of the labels for all the images in the training set, $T^{\neg i}$ is the set of group labels for all the images in the training set except the label for the $i$th image, $\beta_0 = \sum_y^\mathbb{Y} \beta_y$, $\Omega$ is a count matrix, $\Omega_{t, y}$ is the number of images with the group label $t$ and the object class label $y$ as defined in Eq.(\ref{eq:omega}), $\Omega^{\neg i}$ is a count matrix without counting the $\imath$th image.
\begin{equation}\label{eq:omega}
\Omega_{t,y} = \sum_{i=1}^{\Xi}{I(t_i=t \land y_i = y)}
\end{equation}
where $\Xi$ is the total number of labeled training images. 

Eq.(\ref{eq:t}) has two components: (a) The first component is proportional to the number of images in the object class $y_i$ which are assigned into the group $t$, e.g.,  more images in the group $t$ come from the object class $y_i$, higher possibility for the images from the object class $y_i$ to be assigned into the group $t$; (b) The second component is the prediction from the deep networks, e.g., it gives the probability for the image $i$ with the deep representation $x_i$ to be assigned into the group $t$.  The deep representation $x_i$ for the image $i$ may be improved during the iterative process for joint learning of deep networks and our LMM model (tree classifier). New group assignments for the object classes are obtained according to both the improvements of the deep class representations and the improvements of the inter-class similarity relationships for visual hierarchy construction.

The group-object correlation matrix (object-group assignment matrix) $\Psi$ can be estimated automatically and its component $\Psi_{t_i,y_i}$ is obtained as:
\begin{equation}\label{eq:Psi}
\Psi_{t_i,y_i} = \frac{
\Omega_{t_i,y_i} + \beta_{y_i}
}{
\sum_y^\mathbb{Y} \Omega_{t_i,y} + \beta_0
}
\end{equation}

\section{Joint Learning of Tree Classifier, Deep Networks \& Visual Hierarchy Adaptation}\label{training}
In this work, an end-to-end approach is developed to jointly learn: (a) the LMM model (the tree classifier over the visual hierarchy) for recognizing large numbers of object classes hierarchically; (b) the group-object correlation matrix (object-group assignment matrix) $\Psi$ for visual hierarchy adaptation; and (c) the deep networks to extract more discriminative features for image and object class representation. Our joint learning algorithm is illustrated in Algorithm \ref{alg1}.  

\subsection{End-to-end Learning of LMM Model and Deep Networks} 
Joint learning of the LMM model and the deep networks is treated as an issue of Maximum Likelihood Estimation (MLE). The hierarchical recognition error for training the tree classifier over the visual hierarchy can be minimized by maximizing the log likelihood. The set of parameters for the node classifiers at different levels of the visual hierarchy is denoted as $W$, e.g., the tree classifier is represented as a set of node classifiers at different levels of the visual hierarchy. The weights for the deep networks are denoted as $u$, which are used to extract the feature vector $x$ for image and object class representation. The $\log$ likelihood as defined in Eq.(\ref{eq:log}) is used for training the tree classifier over the visual hierarchy:
\begin{equation}\label{eq:log}
\pounds(W,u) = \log \left( \sum_{l=1}^L{\theta_l \sum_t^{\mathbb{T}_l}{  P(t|l,x,w^l_t) P(y|t)}} \right)
\end{equation}
where $w^l_t$ is the parameter of the node classifier for the group $t$ at the $l$th level of the visual hierarchy, $W$ is the set of parameters for all the node classifiers at different levels of the visual hierarchy, $W =\{w^l|^L_{l=1}\}$, $w_l$ is the set of parameters for all the group classifiers at the $l$th level of the visual hierarchy, $w_l =\{w^l_t, t \in \mathbb{T}_l\}$, $\theta_l$ is the leveraging parameter to measure the contributions or effects from the node classifiers at the $l$th level of the visual hierarchy, $\mathbb{T}_l$ is used to note the particular partitioning of all the object classes at the $l$th level of the visual hierarchy (i.e., $N_l$ groups at the $l$th level of the visual hierarchy), $L$ is the depth of the visual hierarchy.

Given the log likelihood (joint objective function) $\pounds(W,u)$, all the parameters $W$ for the node classifiers at different levels of the visual hierarchy and all the parameters (weights) $u$ for the deep networks are learned by using stochastic gradient descent (SGD) \cite{zinkevich-nips2010,bottou-2010,azadi-icml2014,ouyang-icml2013}. By maximizing the log likelihood (joint objective function) $\pounds(W,u)$ as defined in Eq.(\ref{eq:log}), our LMM algorithm can achieve a path-based approach for learning the tree classifier over the visual hierarchy, e.g., learning the set of parameters $W$ for the tree classifier. By back-propagating the gradients $\frac{\partial \pounds(W,u)}{\partial u}$ of the objective function to fine-tune the weights $u$ of the deep networks, our LMM algorithm can provide an end-to-end approach for learning the tree classifier and the deep networks jointly. Through the back-propagation process, our LMM algorithm can: (a) advise some 'common' neurons on the deep networks to learn more discriminative features for supporting more effective separation of the group nodes  at the $l$th level of the visual hierarchy;  and (b) guide some 'specific' neurons to learn more discriminative feature for achieving more accurate recognition of the fine-grained object classes at the bottom level of the visual hierarchy. 

By fixing the group-object correlation matrix $\Psi$ (i.e., fixing the object-group alignments or fixing the tree structure of the visual hierarchy), all the parameters $W$ for the node classifiers at different levels of the visual hierarchy and all the parameters (weights) $u$ for the deep networks are learned by maximizing the log likelihood $\pounds(W,u)$. The parameters of the tree classifier $W$  are updated effectively by back-propagating the gradients $\frac{\partial \pounds(W,u)}{\partial W}$ over the relevant node classifiers on the visual hierarchy. The parameters (weights) for the deep networks are fine-tuned effectively by back-propagating the gradients $\frac{\partial \pounds(W,u)}{\partial u}$=$\frac{\partial \pounds(W,u)}{\partial x}\frac{\partial x}{\partial u}$ of the objective function (log likelihood) over the deep networks. The computation efficiency of our LMM algorithms is  very competitive compared with the traditional $N$-way flat softmax method as shown in Eq. (\ref{eq:complex}).
\begin{equation}
O(d\sum_i^{\log_b^N}{b^i}) = O(d\frac{(1-b^{\log_b^N})}{(1-b)} +Nd) =  O(Nd)
\label{eq:complex}
\end{equation}
where $N$ is used to note the total number of object classes, $d$ is used to note the dimension of the feature vector, $b^i$ is used to note the number of branches for visual hierarchy construction.  

\begin{algorithm}[!t]
 \caption{Our algorithm for jointly learning deep networks, tree classifier and visual hierarchy adaptation}
 \label{alg1} 
\begin{algorithmic}
\State \textbf{Data}: Image set S with $\Xi$ images, Label set Y, Pre-trained visual hierarchy. 
\State Initialize group-object correlation matrix $\Psi$ by using the pre-trained visual hierarchy. 
  \For{epoch = 1, ..., max of iterations} 
      \State update the set of parameters $W, u$ for LMM model (tree classifier) and deep networks by maximizing Eq. (\ref{eq:log}) based on the back-propagated the gradients $\frac{\partial \pounds(W,u)}{\partial W}$, $\frac{\partial \pounds(W,u)}{\partial u} $
  \EndFor 
  \Repeat 
    \For{i = 1, ..., $\Xi$} 
      \State sample the group label for the image $i$ by using Eq. (\ref{eq:t}) 
      \State update the group-object correlation matrix $\Psi$ by using Eq.(\ref{eq:Psi}) 
      \State update the set of parameters $W, u$ for LMM model (tree classifier) and deep networks by maximizing Eq. (\ref{eq:log}) based on the back-propagated the gradients $\frac{\partial \pounds(W,u)}{\partial W}$, $\frac{\partial \pounds(W,u)}{\partial u} $
    \EndFor 
  \Until{converge} 
\end{algorithmic}
\end{algorithm}

\subsection{Back-Propagation Process}
Given the log likelihood (joint objective function) $\pounds(W, u)$, the SGD method \cite{zinkevich-nips2010,bottou-2010,azadi-icml2014,ouyang-icml2013} is used to learn the tree classifier and the deep networks jointly: (a) The gradients of the log likelihood $\frac{\partial \pounds(W,u)}{\partial W}$ are calculated and such gradients are used to update the set of the parameters $W =\{w^l|^L_{l=1}\}$ of the tree classifier; (b)  The gradients of the log likelihood $\frac{\partial \pounds(W,u)}{\partial u}$ are calculated and such gradients are back-propagated to fine-tune the weights $u$ of the deep networks.
\begin{equation}
\frac{\partial \pounds(W,u)}{\partial u} = \frac{\partial \pounds(W,u)}{\partial x}\frac{\partial x}{\partial u}\label{eq:uq}
\end{equation}
By maximizing the log likelihood $\pounds(W,u)$, our LMM algorithm can optimally minimize the hierarchical recognition error effectively, and learn a reliable group-object correlation matrix $\Psi$ to control the inter-level error propagation effectively.

In our LMM model, for each node at the $l$th level of the visual hierarchy, a softmax output is used to estimate the probability $P(t|l,x,w^l_t)$ for the image with deep representation $x$ to be assigned into the group $t$ at the $l$th level of the visual hierarchy. For a given image-class (feature-label) pair ($x_i, y_i$), its group label $t_i$ is defined as the ancestor for its object class $y_i$ on the visual hierarchy:
\begin{equation}
z^{l} = Softmax(w^{l}x_i + b^{l})\label{eq:LMMzl}
\end{equation}
where $w^{l}$ is the set of the parameters for all the node classifiers at the $l$th level of the visual hierarchy, $b^{l}$ is the set of biases. 

\subsubsection{Path-based Training of Tree Classifier}
In our path-based approach (our LMM model) for tree classifier training, the prediction probability $P(y|x,W)$ of the object class $y$ for the given image with deep representation $x$ is obtained as:
\begin{equation}
z = \sum_l^L \theta^l z^l \Psi^l\label{eq:LMMz}
\end{equation}
where $\theta^{l}$ is the leveraging parameter for characterizing the contributions or effects from the node classifiers at the $l$th level of the visual hierarchy, $\Psi^l$ is used to note the group-object correlation matrix $\Psi$ at the $l$th level of the visual hierarchy.  

The hierarchical loss function for path-based training of the tree classifier is defined as the negative of the log likelihood:
\begin{equation}
\pounds(W,u) =  -\log{z_{y_i}}\label{eq:LMMl}
\end{equation}

The gradients of the softmax output for the node classifiers at the $l$th level of the visual hierarchy are obtained as:
\begin{equation}
\frac{\partial \pounds(W,u)}{\partial z^l_t} =\frac{\partial \pounds(W,u)}{\partial z_{y_i}}
\frac{\partial z_{y_i}}{\partial z^l_t} 
 = \theta^l  \Psi^l_{t,{y_i}} \frac{\partial \pounds(W,u)}{\partial z_{y_i}} 
\label{eq:LMMdzl}
\end{equation}
where $\Psi^l_{t,{y_i}}$ is used to note the group-object correlation between the group $t$ at the $l$th level of the visual hierarchy and the object class with the label $y_i$.  Such gradient $\frac{\partial \pounds(W,u)}{\partial z^l_t}$ can be used to measure the effects on the improvements of the parameters of the node classifier for the group $t$ at the $l$th level of the visual hierarchy, and the weights of the deep network which are contributed by the softmax outputs at the $l$th level of the visual hierarchy. In Eq. (\ref{eq:LMMdzl}), $\theta_l$ decides how much effect from the given image ($x_i,y_i$)  can be added to the node classifiers at the $l$th level of the visual hierarchy. $\Psi^l_{t,y}$ decides how much effect from the group-object correlation can be added to the node classifier for the group $t$ at the $l$th level of the visual hierarchy. 

In the higher level of the visual hierarchy,  if there is no adaptation on the object-group assignment matrix $\Psi$, Eq. (\ref{eq:LMMdzl}) can be simplified as Eq. (\ref{eq:LMMdzlh}).
\begin{equation}
\frac{\partial \pounds(W,u)}{\partial z^{l}_{t^l_i}} 
 = \theta^{l} \Psi^l_{{t^l_i},{y_i}} \frac{\partial \pounds(W,u)}{\partial z_{y_i}}  
 =  \frac{\theta^{l}}{C_{t^l_i}} \frac{\partial \pounds(W,u)}{\partial z_{y_i}} 
\label{eq:LMMdzlh}
\end{equation}
where $C_{t^l_i}$ is used to note the number of object classes for the group $t$ at the $l$th level of the visual hierarchy when the given image $i$ is assigned into $t$. 

For the gradients derived from the softmax outputs of the node classifiers at the $l$th level of the visual hierarchy, back-propagation \cite{lecun1998} is used to leverage such gradients to update the parameters of the node classifier for the current node, and thus the corresponding modification on the classifier parameter $\Delta w_{t}^l$ for the current node (i.e., group $t$ at the $l$th level of the visual hierarchy) is defined as:
\begin{equation}
\Delta w_{t}^l = \epsilon \frac{\partial \pounds(W,u)}{\partial w^{l}_{t}}  
 = \epsilon \frac{\theta^{l}}{C_{t^l_i}} \frac{\partial \pounds(W,u)}{\partial z_{y_i}}  \frac{\partial z_{t^l_i}^l}{\partial w^l_{t}} 
\label{eq:LMMdwl}
\end{equation}
where $\epsilon$ is used to note the learning rate. Such gradients for the group $t$ at the $l$th level of the visual hierarchy are further used to update the classifier parameters for the lower-level nodes until the most relevant leaf nodes, which treat the group $t$ as their ancestor on the visual hierarchy.

The gradients of the softmax output for the node classifiers at the bottom level of the visual hierarchy (i.e., for the object classes) are obtained as:
\begin{equation}
\frac{\partial \pounds(W,u)}{\partial z^{l_{bottom}}_{t^{l_{bottom}}_i}} 
 = \theta^{l_{bottom}} \frac{\partial \pounds(W,u)}{\partial z_{y_i}} 
\label{eq:LMMdzlb}
\end{equation}
where $\theta^{l_{bottom}}$ is the leveraging parameter to measure the effects or contributions from the node classifiers at the bottom level of the visual hierarchy, $t^{l_{bottom}}_i$ is used to note the group at the bottom level of the visual hierarchy. At the bottom level of the visual hierarchy, each object class is treated as one single group, thus the group label for the bottom level $t^{l_{bottom}}_i$ is same the object class label $y_i$ for the given image $i$.

Our path-based approach can control the inter-level error propagation effectively: (a) For a given group $t$ at the $l$th level of the visual hierarchy, the gradients of its node classifier as defined in Eq. (\ref{eq:LMMdzl}) are used to update both the classifier parameters for itself and the classifier parameters for the lower-level nodes until the most relevant leaf nodes, which treat the group $t$ as their ancestor on the visual hierarchy; (b) For a given object class $y$ at the bottom level of the visual hierarchy, $\Psi^{l_{bottom}}_{t,y} = 1$, if $t=y$, else $\Psi^{l_{bottom}}_{t,y} = 0$, the gradients of its node classifier as defined in Eq. (\ref{eq:LMMdzl}) are used to update only the classifier parameters for itself.  

\subsubsection{Fine-tuning Network Weights}
The gradients derived from all the softmax outputs at different levels of the tree classifier (our LMM model) are then integrated to update the weights of the deep networks as given  Eq. (\ref{eq:LMMdwx}). 
\begin{equation}
 \frac{\partial \pounds(W,u)}{\partial x_i}  
 =  \sum_l \left \{\frac{\theta^{l}}{C_{t^l_i}} \frac{\partial \pounds(W,u)}{\partial z_{y_i}}  \frac{\partial z_{t^l_i}^l}{\partial x_i} \right \}
\label{eq:LMMdwx}
\end{equation}
where $\sum_l \theta^l = 1$, $\theta^{l}$ is the leveraging parameter to measure the effects or contributions from the node classifiers at the $l$th level of the visual hierarchy. If $\theta^{l_{bottom}} = 1$, our LMM model is exactly same as the traditional $N$-way flat softmax model.

Without using non-overlapping constraint but with adaptation on the group-object correlation matrix $\Psi$, the derivative from our overall model is shown in Eqs. (\ref{eq:LMMdwla}, \ref{eq:LMMdxla}). 
\begin{equation}
\Delta w_{t}^l = \epsilon \sum_{t'}^{\mathcal{T}_l} \left \{ \theta^{l} \Psi^l_{t',y_i} \frac{\partial \pounds(W,u)}{\partial z_{y_i}}  \frac{\partial z_{t'}^l}{\partial w^l_{t}} \right \}
\label{eq:LMMdwla}
\end{equation}
\begin{equation}
\frac{\partial \pounds(W,u)}{\partial x_i}   = \sum_l \sum_{t}^{\mathcal{T}_l} \left \{ \theta^{l} \Psi^l_{{t},{y_i}} \frac{\partial \pounds(W,u)}{\partial z_{y_i}}  \frac{\partial z_t^l}{\partial x_i} \right \}
\label{eq:LMMdxla}
\end{equation}
By adapting the group-object correlation matrix and integrating back-propagation to leverage the gradients of the log likelihood (the objective function for tree classifier training) to fine-tune the weights of our deep networks, our LMM algorithm can provide an unique process for jointly learning:  (a) the deep networks for image and object class representation; (b) the tree classifier (LMM model) for recognizing large numbers of object classes hierarchically; and (c) the visual hierarchy adaptation for achieving more accurate and hierarchical indexing of large numbers of object classes. By jointly learning the visual hierarchy adaptation, the tree classifier and the deep networks in an end-to-end fashion, our LMM algorithm can provide an effective solution for controlling inter-level error propagation effectively and achieve better accuracy rates on large-scale visual recognition.

\subsection{Visual Hierarchy Adaptation}
First, the group-level labels for the images and the group-object correlation matrix (object-group assignment matrix) $\Psi$ are initialized by the pre-trained visual hierarchy as described in Section \ref{LMM}. The group-object correlation matrix $\Psi$ is adapted to the improvements of the outputs of the deep networks as described in Section \ref{tree structure learning}.

Without any constraint, our Bayesian approach may result in an overlapping tree structure (i.e., some uncertain object classes can be assigned into multiple groups simultaneously rather than one single group). In our LMM model, it is really simple to cut the overlapped branches by using a non-overlapping constraint: $\Psi_{t, y} = 0$, where $\Psi_{t,y} \neq \max_t^{\mathbb{T}_l}{\Psi_{t,y}}$.

\subsection{Deep Learning with Regularization}
The process for joint learning of deep networks and tree classifier can also be treated as the process of maximizing a posteriori estimation (MAP). The prior for the set of the parameters $W$ (for all the node classifiers at different levels of the visual hierarchy) is chosen as Gaussian distribution with diagonal covariance.  For example,
\begin{equation}w_i \sim \mathcal{N}\left( 0, \frac{1}{\alpha}I\right),\: w_i \in W\end{equation}
To learn the deep networks and the tree classifier jointly, we can maximize the  $\log$ posterior likelihood: 
\begin{equation}\label{eq:map}
\begin{aligned}
\max_W \{\pounds'(W,u)\} &=\max_W\left\{\log P(y|x,W) + \log P(W) \right\} \\
&=\max_W \left\{\log P(y|x,W)  - \frac{\alpha}{2}\|W\|^2\right\} 
\end{aligned}
\end{equation}
The first term in Eq.(\ref{eq:map}) is  same  as that in Eq.(\ref{eq:log}). The second term  is  L2-norm regularization over the classifier parameter $W$, which is used to control over-fitting and learn more discriminative tree classifier. Eq.(\ref{eq:map}) can be used to replace  Eq.(\ref{eq:log}) in Algorithm\ref{alg1}.

\section{Experimental Results for Algorithm Evaluation}\label{Experiments}
\begin{table}
\caption{\bf The performance comparison between our LMM algorithm and the baseline method. Notations: P: group-object correlation matrix $\Psi$ modification, TC: tree structure updating, w/: with, w/o: without. }
\begin{center}
\begin{tabular}{ l  | c | c | c| c}
\hline
\multirow{2}{*}{Method} & \multicolumn{4}{|c}{Prediction error (Top k)} \\\cline{2-5}
& 1  &2 &5 &10  \\
\hline \hline
  baseline						& 45.13\%	&33.29\% &21.67\% &15.21\% 	\\
  LMM						& 45.05\% 	& 33.30\% 	& 21.64\% 	&15.08\%\\
  LMM w/P, w/o TC	& 44.64\%	& 32.93\%	& \textbf{21.24\%} 	&14.87\% \\
  LMM  w/P,TC 			& \textbf{44.29\%}	& \textbf{32.65\%} 	& 21.26\%
  &\textbf{14.86\%}\\
\hline
\end{tabular}
\end{center}
\label{tab:result on imagenet2012}
\end{table}
Our experiments are carried over two image sets: ImageNet1K and ImageNet10K. ImageNet1K image set contains 1,000 object classes, which are mutual exclusive or overlap but not subsumption in the semantic space. As shown in Fig. \ref{fig:tree1}, the visual hierarchy is pre-trained to organize $1,000$ object classes hierarchically. ImageNet10K image set contains 10,184 image categories, which come from different levels of the concept ontology and some of them could be subsumption of others, e.g., not all of these $10,184$ image categories are semantically atomic (mutually exclusive) because some of them are from the high-level non-leaf nodes of the concept ontology and they are not semantically atomic with others. Thus the concept ontology is incorporated to decompose such high-level image categories (from the non-leaf nodes of the concept ontology) into multiple object classes (at the leaf nodes of the concept ontology), and $7,756$ object classes are finally identified for ImageNet10K image set. As shown in Fig. \ref{fig:tree2}, the visual hierarchy is also pre-trained to organize $7,756$ object classes hierarchically.

We have compared our LMM algorithm with the baseline method \cite{NIPS2012_4824,ding2014theano} and our comparison experiments focus on evaluating multiple factors: (a) whether our LMM algorithm can control the inter-level error propagation more effectively as compared with other baseline methods and achieve better accuracy rates on large-scale visual recognition; (b) whether our LMM algorithm can jointly learn the tree classifier, the deep networks and the visual hierarchy adaptation effectively; and (c) whether our LMM algorithm can achieve higher prediction confidences.

Our experiments are implemented on Theano\cite{2016arXiv160502688short}, with one GPU NVIDIA GTX 980i. In our experiments, the learning rate is set as 0.0001, the dropout rate is set as 0.5 to prevent over-fitting. We take an initial from Theano Alexnet pre-trained \cite{ding2014theano}. 
\begin{figure}
    \centering \includegraphics[width=1.0\linewidth]{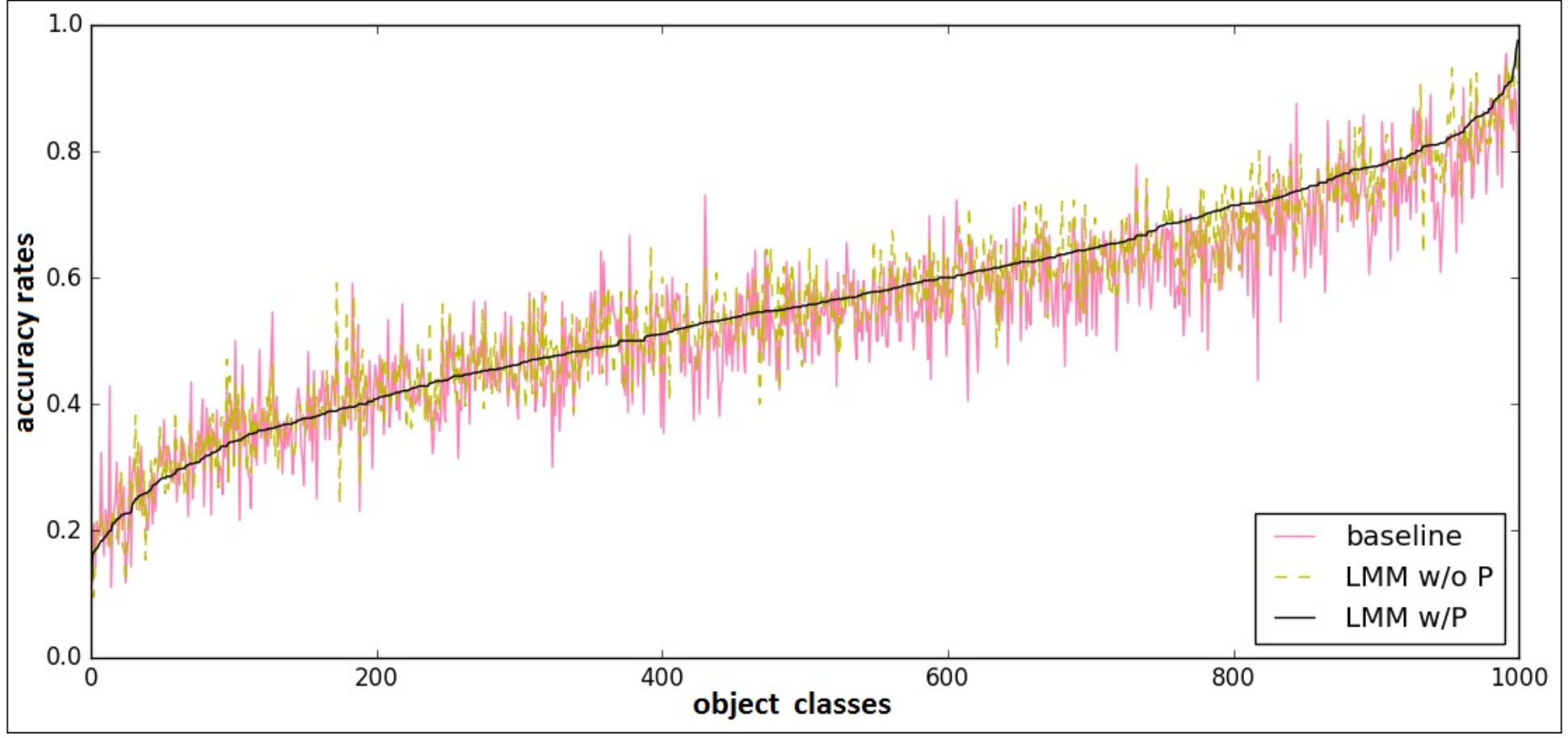}
    \caption{\bf The comparison on the accuracy rates for 1,000 object classes in ImageNet1K image set. }
    \label{fig:avg-comp}
    \vspace*{-0.48cm}
\end{figure}

\subsection{Experimental Results on Imagenet1K}
For the ImageNet1K image set, a deep CNNs with a $N$-way flat softmax classifier is first trained as the baseline \cite{NIPS2012_4824,ding2014theano}. Second, our LMM model (tree classifier) is trained by using a pre-trained visual hierarchy (without modifying the group-object correlation matrix $\Psi$ along the time). Third, our LMM model is trained with modifications on the group-object correlation matrix $\Psi$.  Finally,  our LMM model is trained by jointly learning the deep networks, the tree classifier, and the visual hierarchy adaptation in an end-to-end fashion. The experimental results on the average accuracy rates are shown in Table \ref{tab:result on imagenet2012}, one can easily observe that: (a) our LMM algorithm can successfully outperform the baseline method; (b) When our LMM algorithm jointly learns the deep networks, the tree classifier, and the visual hierarchy adaptation in an end-to-end fashion, it can achieve the best performance.  

\begin{figure*}
    \centering \includegraphics[width=0.88\linewidth]{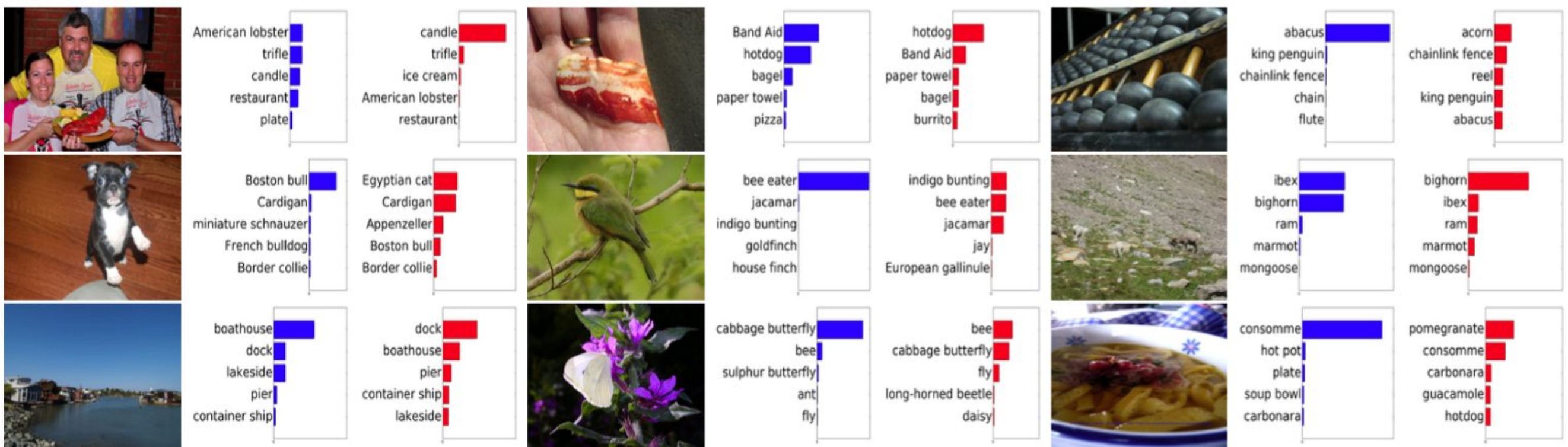}
    \caption{\bf Some experimental results on visual recognition: (a) test images; (b) recognition results and confidence socires from our LMM model (in blue color); (c) recognition results and confidence scores from baseline method (in red color). }
    \label{fig:exp}
    \vspace*{-0.48cm}
\end{figure*}

\begin{figure*}
    \centering \includegraphics[width=0.88\linewidth]{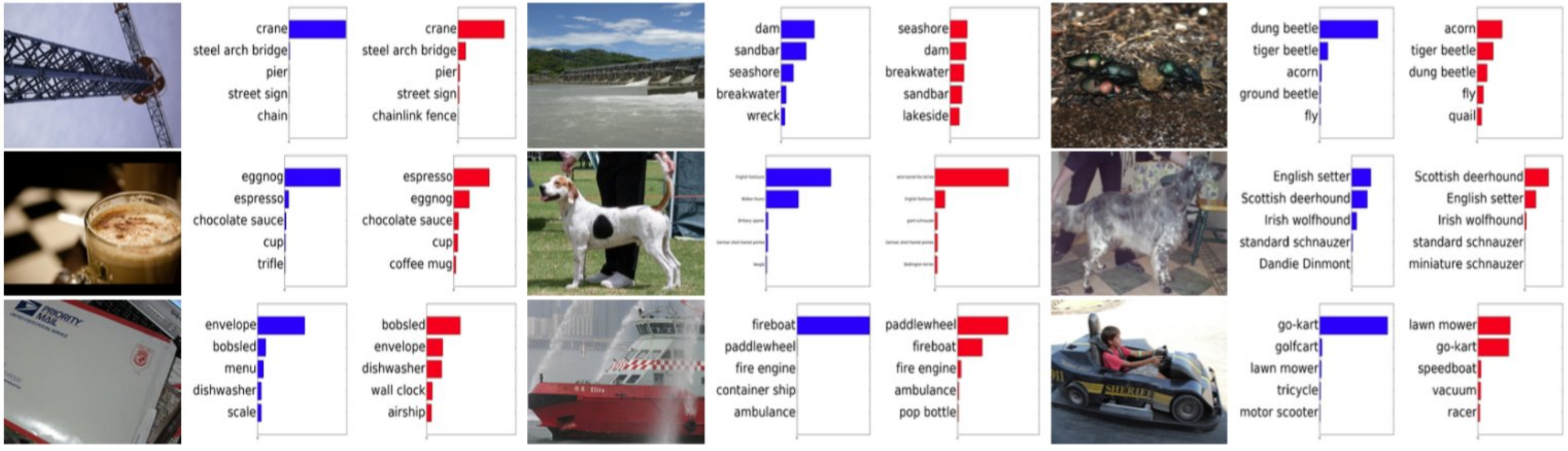}
    \caption{\bf Some experimental results on visual recognition: (a) test images; (b) recognition results and confidence socires from our LMM model (in blue color); (c) recognition results and confidence scores from baseline method (in red color). }
    \label{fig:exp2}
\end{figure*}

\begin{figure*}
    \centering \includegraphics[width=0.88\linewidth]{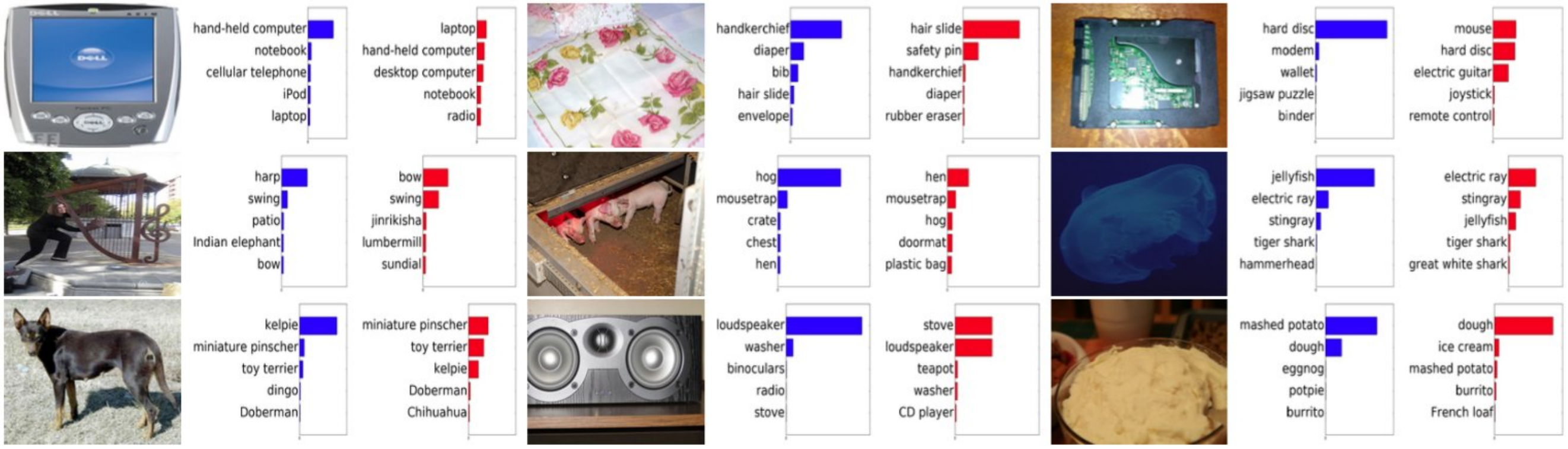}
    \caption{\bf Some experimental results on visual recognition: (a) test images; (b) recognition results and confidence socires from our LMM model (in blue color); (c) recognition results and confidence scores from baseline method (in red color). }
    \label{fig:exp3}
\end{figure*}

The comparisons on the accuracy rates are shown in Fig.  \ref{fig:avg-comp}, for all these 1,000 object classes in the ImageNet1K image set, 56\% object classes have obtained better accuracy rates by using our LMM model, 19\% object classes remain no obvious changes on their accuracy rates, 25\% object classes loss more than 1\%  accuracy rates. Some visual recognition examples and their confidence scores are shown in Fig. \ref{fig:exp}, Fig. \ref{fig:exp2} and Fig. \ref{fig:exp3}, one can easily observe that our LMM algorithm can achieve more accurate recognition with higher confidence scores. 

\subsubsection{Accuracy Rates vs. Learning Complexities}
As we mentioned before, one goal for embedding the visual hierarchy with the deep networks is to deal with the issues of strong inter-class visual similarities and diverse learning complexities for large-scale visual recognition. Thus it is very attractive to develop new algorithms to: (a) separate such visually-similar object classes with similar learning complexities from others; and (b) train the inter-related classifiers for such visually-similar object classes jointly. By integrating the visual hierarchy to assign the visually-similar object classes into the same group, our LMM algorithm can learn their inter-related classifiers jointly. Because such visually-similar object classes share similar learning complexities, the gradients of their joint objective function are more uniform and homogeneous, so that the back-propagation process can easily stick on reaching a global optimum effectively. As a result, our LMM algorithm can achieve higher accuracy rates on distinguishing such visually-similar object classes which are typically hard to be separated. 
\begin{table*}
\caption{\bf Reassignment of the object classes for group No.4: Object classes in red are initially assigned into group No.4 by the pre-trained visual hierarchy, object classes in blue box are new ones which are re-assigned into group No.4 because of visual hierarchy adaptation.  }
\begin{center}
\begin{tabular}{c}
\includegraphics[width=\linewidth]{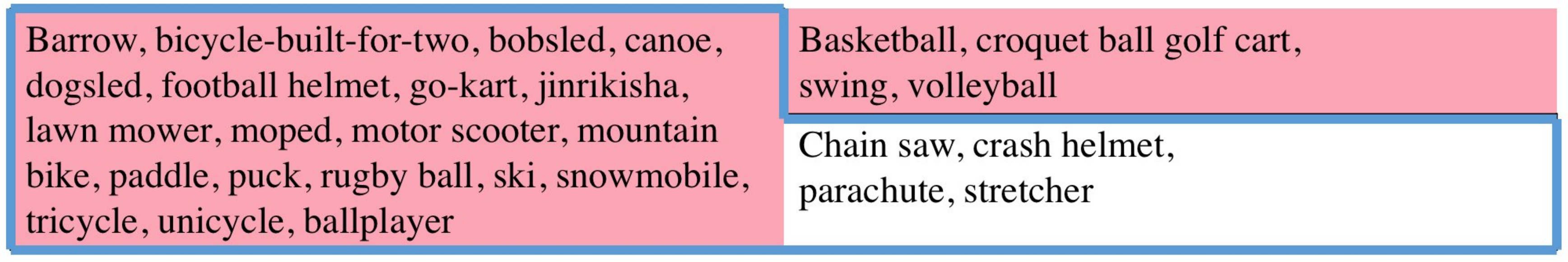}
\end{tabular}
\end{center}
\label{tab:tc4w}
\end{table*}
\begin{table*}
\caption{\bf The lists of object classes with increased and decreased accuracy rates when more representative deep networks are learned along the time.}
\begin{center}
\begin{tabular}{ c}
\includegraphics[width=\textwidth]{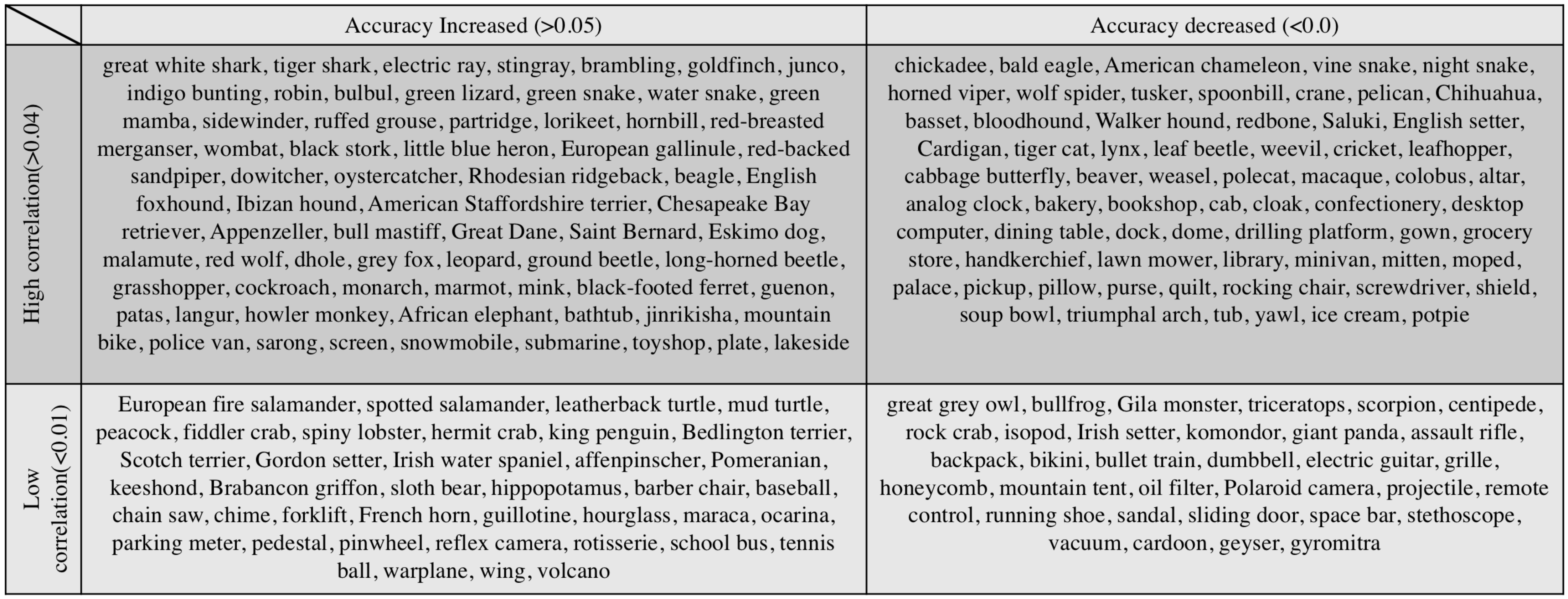}
\end{tabular}
\end{center}
\label{tab:a_objectlist}
\end{table*}
\begin{figure*}
    \centering
  \includegraphics[width=1.0\textwidth]{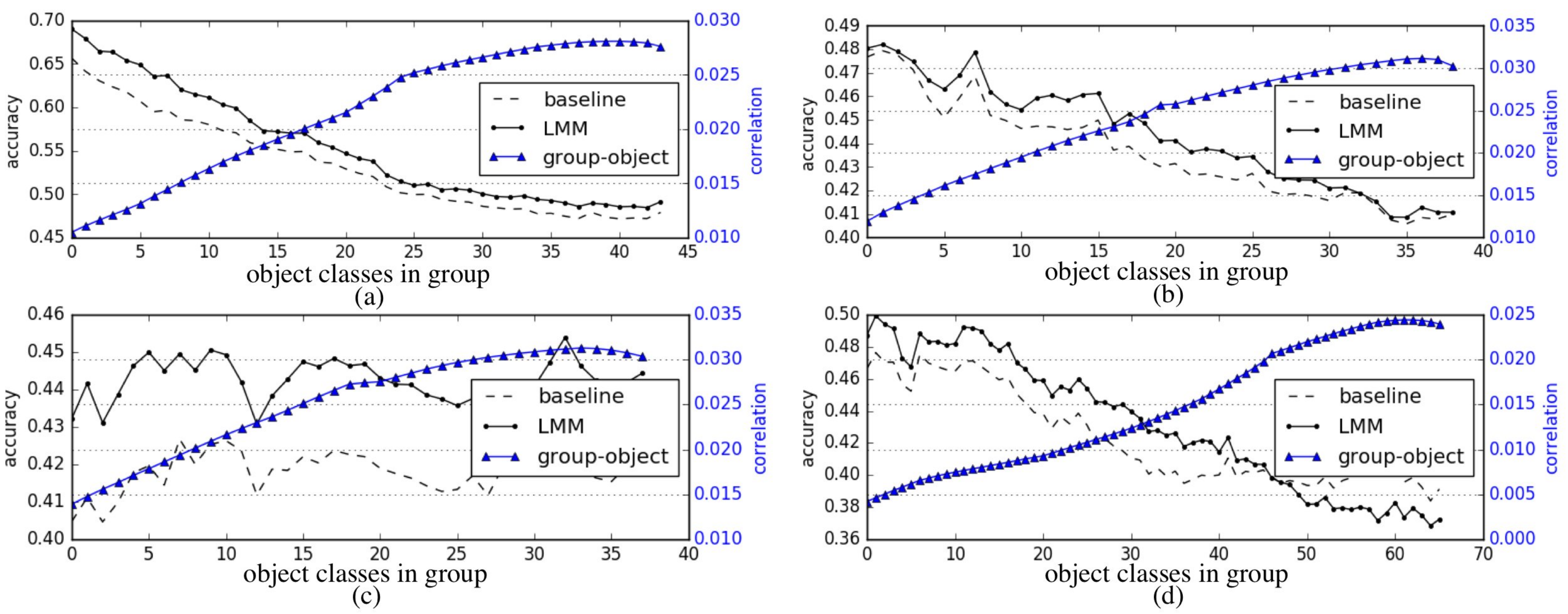}
    \caption{\bf The experimental results for 4 groups of object classes on their correspondences between the accuracy rates for object recognition and the strengths of the group-object correlations.  }
    \label{fig:o_leveraging}
\end{figure*}

Because the deep networks and the tree classifier may not be discriminative enough at the beginning or the pre-trained visual hierarchy may make wrong assignments for some object classes, our LMM model may have low accuracy rates for these object classes which are initially assigned into wrong groups because their deep class representations are not discriminative enough at the beginning. After few iterations, the accuracy rates for these object classes can be improved significantly when they are finally re-assigned into their best-matching groups correctly. The reason for such improvements is that our LMM model can jointly learn the deep networks, the tree classifier and the visual hierarchy (i.e., the object-group assignment matrix), thus they can be improved simultaneously, e.g., more accurate deep class representations (more representative deep networks) can result in more accurate assignments of object classes (more accurate visual hierarchy) and learn more discriminative tree classifier. The names for some of those object classes with improved accuracy rates are listed in Table \ref{tab:a_objectlist}.

In Fig \ref{fig:o_leveraging}, we illustrate our experimental results for 4 groups with different numbers of visually-similar object classes. From these experimental results, one can observe that there have good correspondences between the accuracy rates for object recognition and the strengths of the group-object correlations. From these experimental results, one can observe that: (a) The object classes, which have large values (strengths) of group-object correlations at the beginning, may have small improvements on their accuracy rates along the time because they have already been assigned into their best-matching groups correctly; (b) The object classes, which have low values (strengths) of group-object correlations at the beginning, may have big improvements on their accuracy rates along the time when more representative deep networks are learned and they are finally re-assigned into their best-matching groups correctly. 

The reasons are that: (a) The object classes, which have large values (strengths) of group-object correlations at the beginning, have already been assigned into their best-matching groups correctly by the pre-trained visual hierarchy, and the deep representations for those object classes have already been exploited for training the group-level classifiers effectively and thus they may have less contributions on improving the accuracy rates at the group level; (b) The object classes, which have low values (strengths) of group-object correlations at the beginning, may initially be assigned into wrong groups by the pre-trained visual hierarchy and can be re-assigned into their best-matching groups correctly because of visual hierarchy adaptation, and thus those re-assigned object classes may have more contributions on improving the group-level classification performance along the time. Thus the object classes, which have low values (strengths) of group-object correlations at the beginning, could have significant improvement on the accuracy rates at the object class level. 
 
Overall, by leveraging the visual hierarchy to assign the visually-similar object classes with similar learning complexities into the same group and learn their inter-related classifiers jointly, the gradients of their joint objective function are more uniform and homogeneous, thus our LMM algorithm can obtain global optimum effectively and result in more discriminative tree classifier for large-scale visual recognition.

\subsubsection{Prediction Confidences}
For large-scale visual recognition application, it is also very important to evaluate the confidences of the predictions for object recognition. As shown in  Fig. \ref{fig:exp}, Fig. \ref{fig:exp2} and Fig. \ref{fig:exp3}, by assigning the visually-similar object classes with similar learning complexities into the same group and learning their inter-related classifiers jointly, our LMM model can obtain higher prediction confidence scores as compared with the baseline method. 
\begin{figure}
    \centering \includegraphics[width=1.0\linewidth]{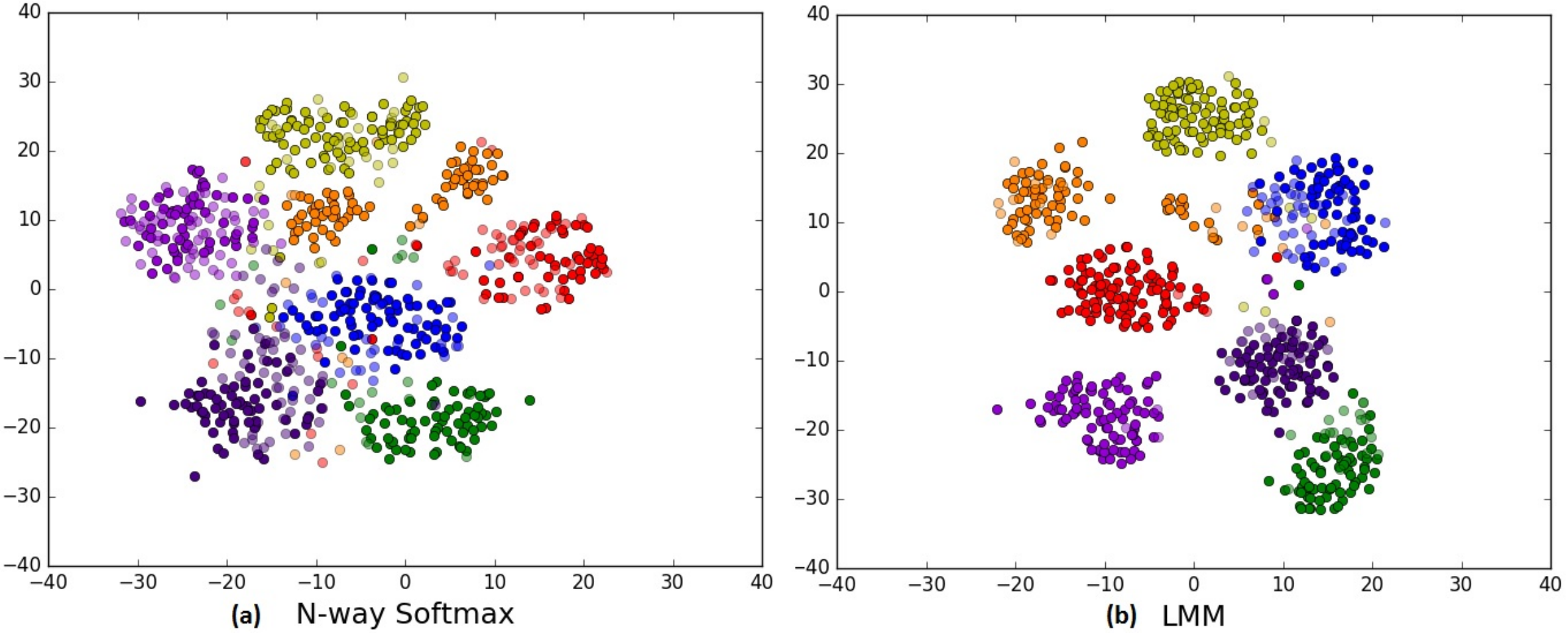}
    \caption{\bf The comparison on the inter-class separability for the visually-similar object classes in the same group. }
    \label{fig:visualization}
    \vspace*{-0.48cm}
\end{figure}

\subsubsection{Visual Hierarchy Adaptation and Object-Group Reassignment}
The effects of visual hierarchy adaptation are evaluated by comparing multiple approaches: (1) our LMM model with an initial group-object correlation matrix $\Psi$ (provided by a pre-trained visual hierarchy); (2) our LMM model with modification of group-object correlation matrix via visual hierarchy adaptation; and (3) our LMM model with modification of group-object correlation matrix and non-overlapping constraint. The names of object classes in that group are shown in Table \ref{tab:tc4w}, the names of the object classes which are initially assigned into that group are  listed in red background in Table \ref{tab:tc4w}. The re-assignments of the object classes from our visual hierarchy adaptation method are listed in Table \ref{tab:tc4w}. The names listed in the blue box are the new object classes which are re-assigned into this particular group and only 20\% object classes are re-assigned because of visual hierarchy adaptation, thus the pre-trained visual hierarchy can achieve reasonable performance on assigning the visually-similar object classes with similar learning complexities into the same group \cite{fan2015,fan2017}. 

\subsubsection{Inter-Class Separability Enhancement}
By assigning the visually-similar atomic object classes with similar learning complexities into the same group, as shown in Fig. \ref{fig:visualization}, our LMM algorithm can significantly enhance their inter-class separability by focusing on learning more discriminative deep representations and node classifiers to enlarge their inter-class margins. By assigning the visually-similar atomic object classes with similar learning complexities into the same group, the gradients of their joint objective function are more uniform and homogeneous, thus our LMM algorithm can obtain global optimum effectively and enlarge their inter-class margins significantly. 
\begin{figure}
    \centering \includegraphics[width=1.0\linewidth]{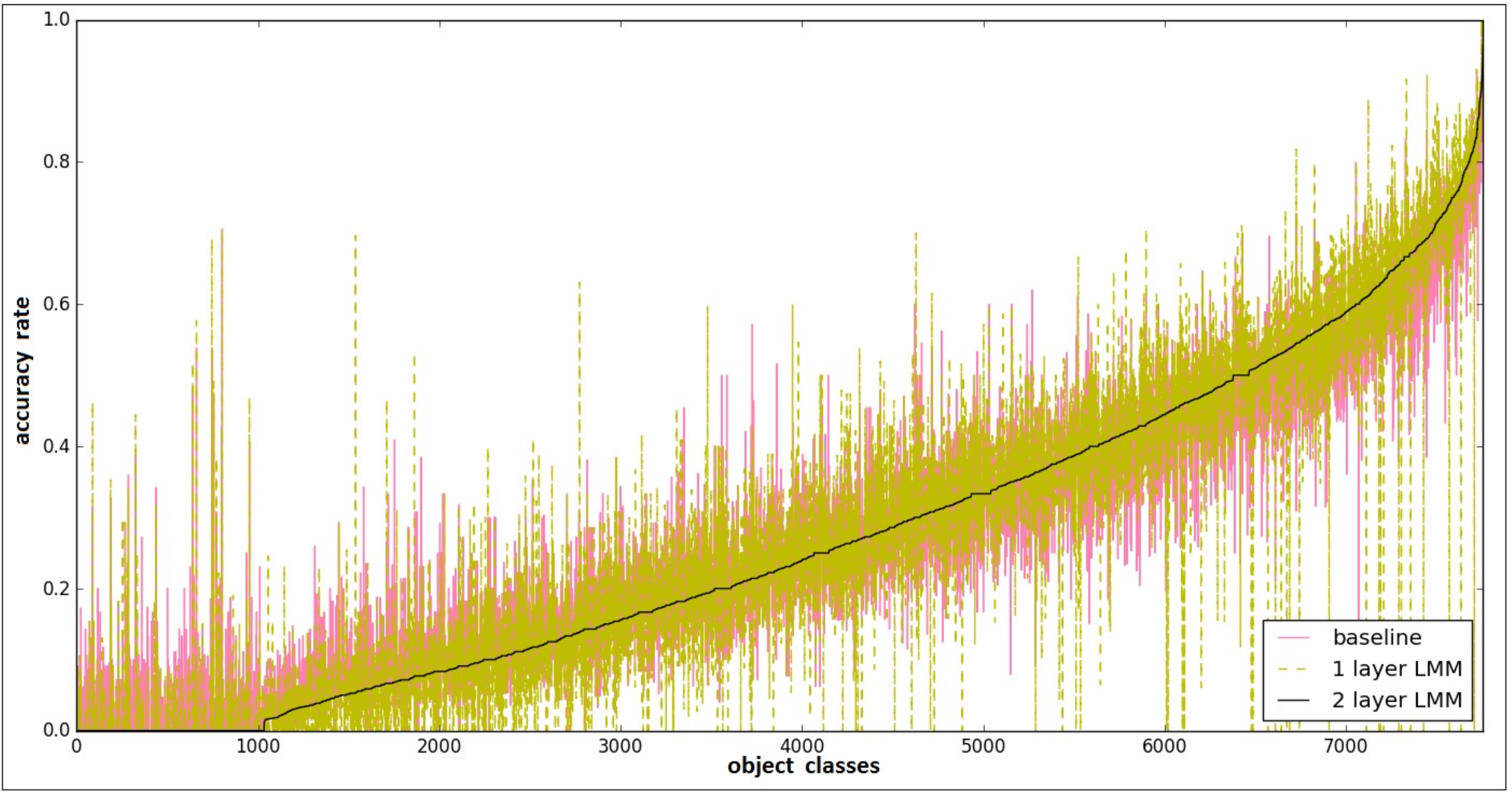}
    \caption{\bf The comparison on the accuracy rates for all the object classes in ImageNet10K image set. }
    \label{fig:avg-comp2}
    \vspace*{-0.48cm}
\end{figure}

\subsection{Experimental Results on Imagenet10K}
For the ImageNet10K image set, we take the parameters from the deep networks, which are already trained for the Imagenet1K image set, as the initials. We then use the images from ImageNet10K to train our LMM model. We have evaluated several kinds of initials, such as the baseline method \cite{NIPS2012_4824,ding2014theano} and our LMM model. We finally find that taking the parameters in our LMM model (which is already trained for Imagenet1K image set) to initialize our deep networks for ImageNet10K image set can allow us to achieve the best performance. 

Because the training cost for the ImageNet10K image set is very high, we only perform our LMM method over one visual hierarchy whose number of branch at each node is 100 (i.e., each parent node contains 100 sibling child nodes). We have also compared our LMM algorithm over the ImageNet10K image set under the following settings: (a) $N$-way flat softmax classifier \cite{NIPS2012_4824,ding2014theano}. (b) One-level LMM model: each group contains only one object class. The group-object correlation matrix is same as the confusion matrix between all the object classes, $\Psi \in R^{N \times N}$. (c) Two-level LMM model: one level for coarse-grained groups and one level for the visually-similar object classes. The experimental results on average accuracy rates are listed on Table \ref{tab:10k}, one can observe that our LMM algorithm can successfully outperform the $N$-way flat softmax classifier. The comparisons on the accuracy rates are shown in Fig.  \ref{fig:avg-comp2}, for all these object classes in the ImageNet10K image set, 68\% object classes have obtained better accuracy rates by using our LMM model, 19\% object classes remain no obvious changes on their accuracy rates, 13\% object classes loss more than 1\%  accuracy rates. 


\begin{table}
\caption{\bf The performance comparison between multiple approaches over the ImageNet10K image set.}
\begin{center}
\begin{tabular}{ l  | c | c | c| c}
\hline
\multirow{2}{*}{Method} & \multicolumn{4}{|c}{Prediction error (Top k)} \\\cline{2-5}
& 1  &2 &5 &10 \\
\hline \hline
  $N$-way Softmax						& 70.30\%	&60.33\% &47.99\% &39.20\% 	\\
  one-level LMM						& 69.60\% 	& \textbf{59.13\%} 	& \textbf{46.87\%} 	&38.82\%\\
  two-level LMM & \textbf{69.50\%}	& 59.39\%	& 46.88\% 	&\textbf{38.07\%} 	\\
\hline
\end{tabular}
\end{center}\label{tab:10k}
\end{table}

By integrating the visual hierarchy to assign the visually-similar object classes into the same group,  such visually-similar object classes in the same group may share similar learning complexities, thus the gradients of their joint objective function are more uniform and homogeneous, so that our LMM algorithm can easily stick on reaching a global optimum effectively and achieve higher accuracy rates on large-scale visual recognition.

\section{Conclusions and Future Works}\label{Conclusion}
A level-wise mixture model (LMM) is developed in this paper to boost the performance of large-scale visual recognition. Our LMM model can provide an end-to-end approach to jointly learn the deep networks for image and object class representation, the tree classifier for recognizing large numbers of object classes hierarchically and the visual hierarchy adaptation for achieving more accurate and hierarchical indexing of large numbers of object classes, thus our LMM algorithm can also provide an effective approach for controlling inter-level error propagation effectively and achieve better accuracy rates on large-scale visual recognition. By seamlessly integrating two divide-and-conquer approaches (deep learning and hierarchical visual recognition), we have found that these two approaches can benefit from each other to exploit better solutions for large-scale visual recognition. Our experimental results on ImageNet1K and ImageNet10K image sets have demonstrated that our LMM algorithm can achieve competitive results on the accuracy rates as compared with the baseline.
 
Many other deep networks, such as VGG \cite{vgg2015}, GoogleNets \cite{googlenet2015}, Resinet \cite{resnet2015}, have successfully designed to recognize 1,000 object classes. Thus it is very attractive to leverage these successful designs of deep networks \cite{vgg2015,googlenet2015,resnet2015} to configure our deep networks and evaluate the performances of our LMM  algorithm when different types of deep networks are used. Because these complex deep networks \cite{vgg2015,googlenet2015,resnet2015} have achieved better performance than AlexNet \cite{NIPS2012_4824,jia2014,donahue2014} used in this paper, we can expect that using these complex deep networks can allow our LMM algorithm to achieve higher accuracy rates on large-scale visual recognition. 

\ifCLASSOPTIONcaptionsoff
  \newpage
\fi



\bibliographystyle{IEEEtran}
\bibliography{zhao2017-3}

\begin{IEEEbiography}[{\includegraphics[width=1.2in,height=1.25in,clip,keepaspectratio]
{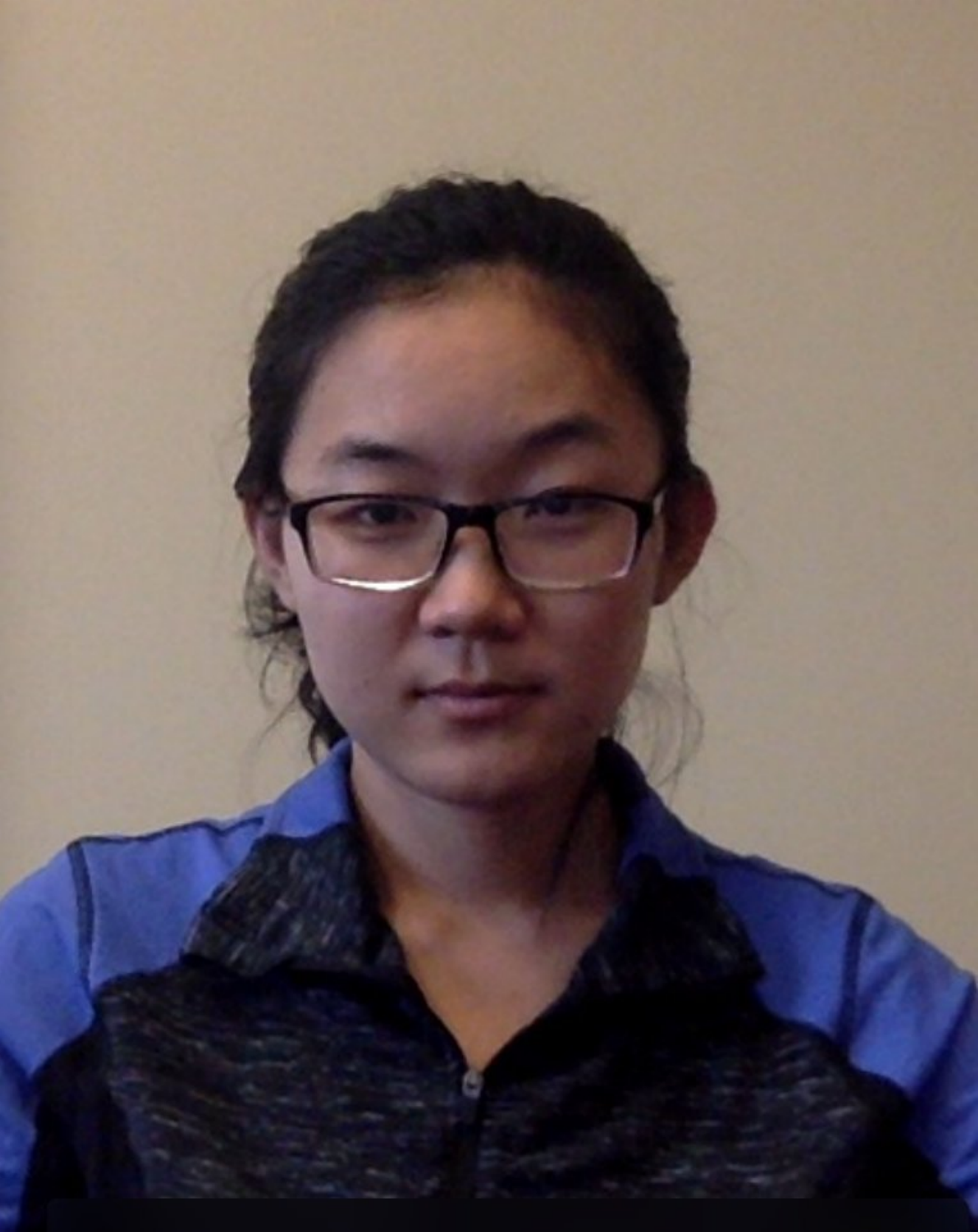}}]{Tianyi Zhao} received her BS degree in software  engineering from Xiamen University in 2015. She is currently pursuing her PhD degree on Computer Science at UNC-Charlotte. Her research interests include semantic image/video classification and retrieval, statistical machine learning, and large-scale visual recognition.
\end{IEEEbiography}
\begin{IEEEbiography}[{\includegraphics[width=1.2in,height=1.25in,clip,keepaspectratio]
{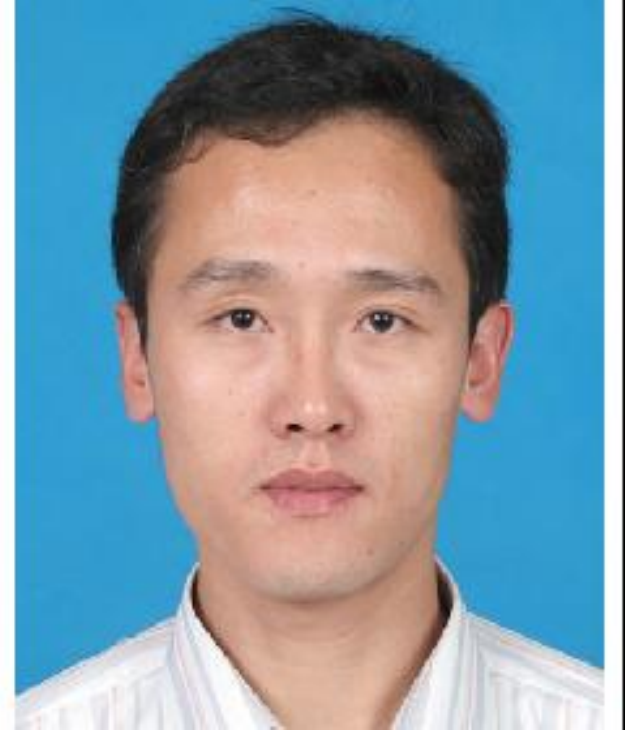}}]{Baopeng Zhang} received his Ph.D. degree in computer science from Tsinghua University in 2008. He is currently an assistant professor in the School of Computer and Information Technology, Beijing Jiaotong University, China. He was a visiting scholar at UNC-Charlotte from 2015-2016. His research interests include semantic image/video classification and retrieval, statistical machine learning, large-scale semantic data management and analysis, and image privacy protection.
\end{IEEEbiography}
\begin{IEEEbiography}[{\includegraphics[width=1.2in,height=1.25in,clip,keepaspectratio]
{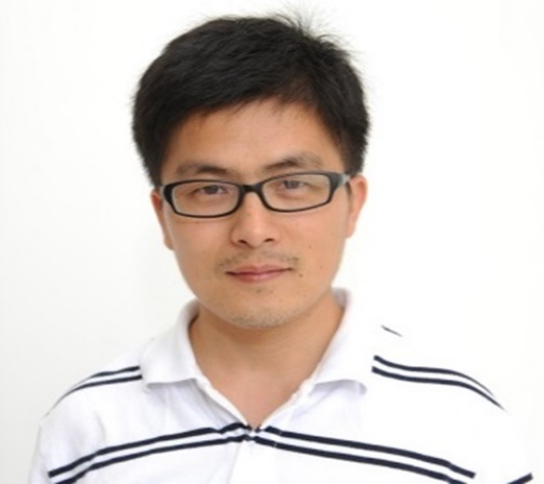}}]{Wei Zhang} received the BA and MA degrees in economics and the PhD degree in computer science from Fudan University, Shanghai, China, in 2000, 2003, and 2008, respectively. He is currently an associate professor in the School of Computer Science, Fudan University. He is now a visiting scholar at the computer science department in University of North Carolina at Charlotte. His current research interests include machine learning, computer vision, and deep neural network.
\end{IEEEbiography}
\begin{IEEEbiography}[{\includegraphics[width=1.1in,height=1.5in,clip,keepaspectratio]
{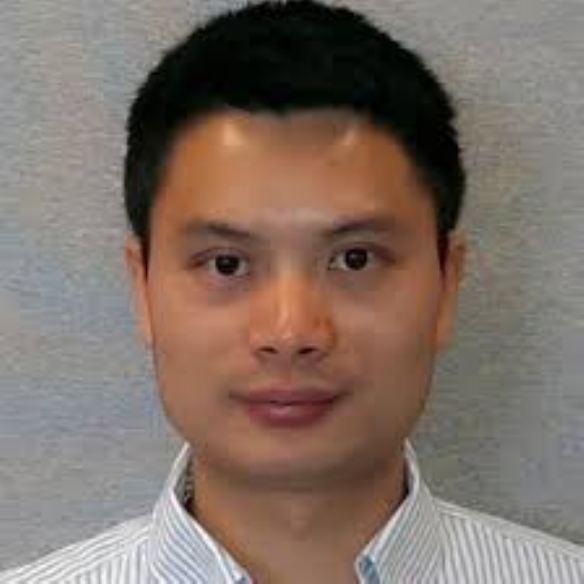}}]{Ning Zhou} received his Ph.D. degree in computer science from UNC-Charlotte in 2013. He is currently a research scientist at Microsoft. His research interests include semantic image/video classification and retrieval, statistical machine learning, large-scale semantic data management and analysis.
\end{IEEEbiography}
\begin{IEEEbiography}[{\includegraphics[width=1.2in,height=1.25in,clip,keepaspectratio]
{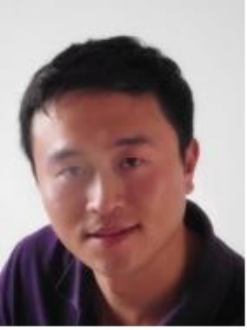}}]{Jun Yu} (M’13) received his BEng and PhD from Zhejiang University, Zhejiang, China. He is currently a Professor with the School of Computer Science and Technology, Hangzhou Dianzi University. He was an Associate Professor with School of Information Science and Technology, Xiamen University. From 2009 to 2011, he worked in Singapore Nanyang Technological University. From 2012-2013, he was a visiting researcher in Microsoft Research Asia (MSRA). He was a short-term visiting scholar at UNC-Charlotte. Over the past years, his research interests include multimedia analysis, machine learning, image processing and image privacy protection. He has authored and co-authored more than 50 scientific articles. He has (co-)chaired for several special sessions, invited sessions, and workshops. He served as a program committee member or reviewer top conferences and prestigious journals. He is a Professional Member of the IEEE, ACM and CCF.
\end{IEEEbiography}
\begin{IEEEbiography}[{\includegraphics[width=1.2in,height=1.25in,clip,keepaspectratio]
{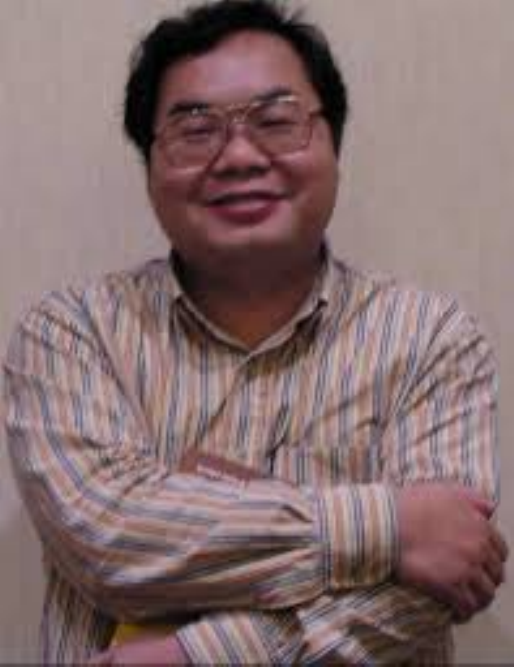}}]
{Jianping Fan} is a professor at UNC-Charlotte. He received his MS degree in theory physics from Northwestern
University, Xian, China in 1994 and his PhD degree in optical storage and computer science from Shanghai
Institute of Optics and Fine Mechanics, Chinese Academy of Sciences, Shanghai, China, in 1997. He was a Postdoc Researcher at Fudan University, Shanghai, China, during 1997-1998. From 1998 to 1999, he was a Researcher
with Japan Society of Promotion of Science (JSPS), Department of Information System Engineering, Osaka
University, Osaka, Japan. From 1999 to 2001, he was a Postdoc Researcher in the Department of Computer
Science, Purdue University, West Lafayette, IN. His research interests include image/video privacy protection, automatic image/video understanding, and large-scale deep learning.
\end{IEEEbiography}

\end{document}